\documentclass{article}

\usepackage[preprint]{neurips_2026}

\usepackage[utf8]{inputenc} 
\usepackage[T1]{fontenc}    
\usepackage{hyperref}       
\usepackage{url}            
\usepackage{booktabs}       
\usepackage{amsfonts}       
\usepackage{nicefrac}       
\usepackage{microtype}      
\usepackage{xcolor}         
\usepackage{xspace}
\usepackage{amsmath}
\usepackage{multirow}
\usepackage{colortbl} 
\usepackage{graphicx}
\usepackage{makecell}
\usepackage{subcaption}
\usepackage{graphicx}
\usepackage{float}

\newcommand{\Method}{CODA\xspace}
\usepackage{enumitem}

\usepackage{wrapfig}

\title{CODA: Cohort- and Drift-aware Foundation Model for Multimodal Clinical Reasoning}

\author{%
  Changshuo Liu\\
  School of Computing\\
  National University of Singapore
\And
  Jiaqi Zhu\thanks{Corresponding author: Jiaqi Zhu (\texttt{jiaqi77@nus.edu.sg})}\\
  School of Computing\\
  National University of Singapore
\And
  Wenqiao Zhang\\
  School of Software Technology\\
  Zhejiang University
\And
  Xiaokui Xiao\\
  School of Computing\\
  National University of Singapore
\And
  Beng Chin Ooi\\
  School of Software Technology\\
  Zhejiang University
}

\begin{document}

\maketitle

\begin{abstract}
  Recent advances in multimodal foundation models have demonstrated strong performance across diverse clinical tasks. However, existing approaches predominantly operate at the individual sample level and fail to model the interplay between population-level cohort patterns and patient-specific variation that underlie real-world clinical reasoning. In practice, clinicians routinely reason through patient cohorts sharing similar demographic, physiological, or pathological characteristics, while accounting for patient-specific deviations from the group prototype. Motivated by this gap, we propose \Method, a cohort- and drift-aware foundation model for multimodal clinical reasoning. \Method explicitly integrates cohort structure into the modeling pipeline through two complementary mechanisms: cohort-aware population stratification and drift-aware individual adaptation. In particular, we leverage reinforcement learning to discover clinically meaningful patient cohorts, and further model each patient's deviation from cohort prototypes as a structured drift signal encoding fine-grained individual heterogeneity. These cohort- and drift-aware representations are incorporated into the foundation model to enable reasoning that is simultaneously informed by population-level patterns and sensitive to individual-level variation. Extensive experiments on real-world electronic health record (EHR) benchmarks demonstrate that \Method achieves consistent effectiveness and strong generalizability across diverse clinical tasks, including closed-QA, open-QA, and report generation.
\end{abstract}

\section{Introduction}
\label{sec:introduction}
Generative AI has rapidly evolved from text-only large language models (LLMs)~\cite{brown2020language,openai202334} to large vision–language models (LVLMs)~\cite{radford2021learning}, enabling unified reasoning over both textual and visual modalities. This paradigm shift is particularly consequential in the medical domain, where clinical decision-making inherently relies on heterogeneous data sources, including medical images, EHR data, and free-text reports~\cite{chen2025generative,ye2025multimodal,kalpelbe2025vision,zhu2026logictree}. By integrating these modalities within a single framework, recent approaches have demonstrated strong capabilities in medical question answering~\cite{li2023llava}, multimodal understanding~\cite{lin2025healthgpt}, and medical decision support~\cite{qiu2024llm}. These advances have driven notable progress in applications such as medical knowledge acquisition and education, treatment planning, and patient outcome prediction~\cite{singhal2025toward,zhang2024generalist}.

Despite these advances, existing medical foundation models~\cite{moor2023med, chen2024huatuogpt, sellergren2025medgemma} are developed in a task-centric manner, where reasoning is performed on individual queries with limited consideration of broader clinical context.
However, clinicians rarely reason about a patient in isolation. Instead, they implicitly compare the current case with previously encountered patients, drawing on experience with recurring patterns across similar populations~\cite{berner2007clinical}. Without this kind of contextual grounding, medical foundation models struggle to capture meaningful clinical variation, often leading to fragmented or overly simplified reasoning.

\begin{figure}[t]
\begin{center}
\includegraphics[width=0.9\textwidth]{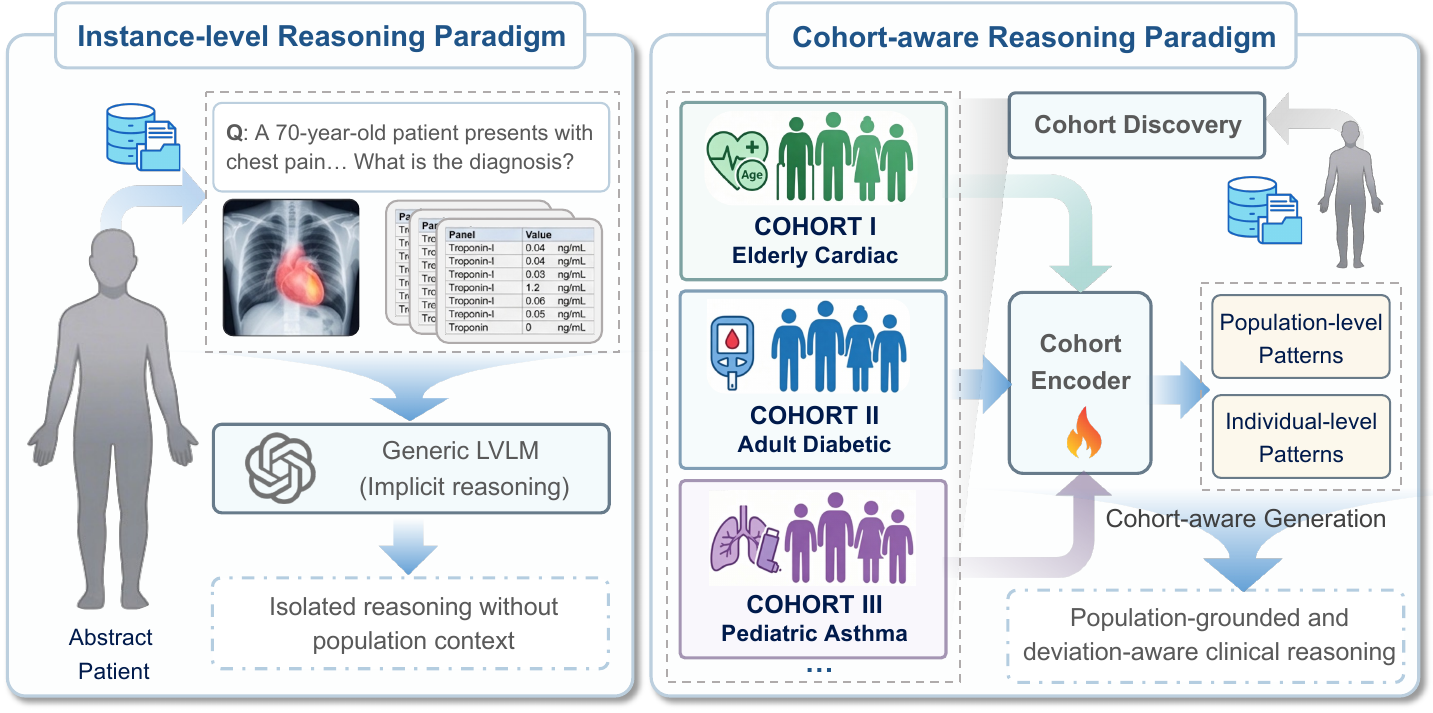}
        \caption{Rethinking Clinical Reasoning: From instance-level predictions to cohort-aware modeling.}
        \label{fig:intro_background}
\end{center}
\vskip -0.2in
\end{figure}


A natural way to incorporate such context is through patient cohorts~\cite{wang2020cohort}. By grouping patients with similar clinical profiles, such as demographics or specific health conditions, cohorts help guide how individual cases are interpreted and managed.
Figure~\ref{fig:intro_background} illustrates the transition from conventional instance-level reasoning to cohort-aware clinical reasoning.
Take the age cohorts as an example, a sudden episode of chest pain may be treated as a potential medical emergency for elderly patients with a history of heart disease, often requiring immediate hospital evaluation, whereas the same symptoms for young and healthy individuals may be attributed to benign causes such as muscle strain or stress.
This type of cohort-based reasoning allows clinical decisions to be grounded in population-level patterns.

Furthermore, relying on cohort structure alone is insufficient to support accurate clinical decisions. Even within the same cohort, patients may differ in ways that lead to different clinical judgments.
For example, two elderly patients with similar cardiovascular risk factors may both have chest pain, yet one may show stable vital signs and mild symptoms that allow for conservative monitoring, while the other may exhibit severe pain that demands immediate intervention.
Such differences in symptoms and clinical signals often determine the final intervention, yet are not well captured by coarse cohort patterns.
As a result, effective clinical decision-making requires jointly modeling cohort structure and individual heterogeneity, a capability that remains underexplored in existing approaches.

To this end, we propose \Method\footnote{Please refer to Appendix~\ref{sec:appendix-related-work} for a discussion of related work.}, a \textbf{CO}hort- and \textbf{D}rift-\textbf{A}ware foundation model that explicitly characterizes both cohort structure and individual heterogeneity.
Our framework consists of two key components.
First, we formulate cohort construction as a reinforcement learning process that automatically partitions patients into clinically meaningful groups. This process builds a cohort tree, where each internal node corresponds to a feature-based status and action, and each leaf represents a cohort.
A cohort encoder is then employed to encode these cohorts, and the resulting embeddings are injected into the model input to provide cohort-level context.
Second, we move beyond static cohorts and introduce a continuously evolving cohort prototype space, where cohort representations are dynamically updated.
For each patient, we further compute a cohort-drift signal that captures how the individual deviates from the corresponding cohorts.
This signal reflects patient-specific characteristics that are not explained by shared cohort patterns, and is incorporated into the model to refine its reasoning at the individual level. These cohort-aware and drift-aware representations enable more precise and context-aware clinical reasoning.
We summarize our main contributions as follows:
\begin{itemize}[itemsep=0mm,leftmargin=4mm]
\vspace{-2mm}

\item We propose a cohort- and drift-aware foundation model for multimodal clinical reasoning that jointly incorporates population-level cohort structure and patient-specific heterogeneity.

\item We present a reinforcement learning-based cohort discovery and dynamic prototype updates strategy, enabling adaptive and data-driven cohort modeling.

\item We explicitly model the individual deviations from cohort prototypes, and incorporate such signals into the backbone through drift-conditioned adapters.

\item Extensive experiments demonstrate that \Method consistently outperforms both general-purpose and medical-specific baselines across closed-QA, open-QA, and report generation tasks on diverse real-world EHR datasets, highlighting its effectiveness in enhancing clinical reasoning.
\end{itemize}

\section{Preliminary}
\label{sec:preliminary}

\subsection{Multimodal Clinical Data Representation in Foundation Models}

Modern foundation models encode heterogeneous clinical data by projecting multimodal inputs into a unified token embedding space. 
Given the input $\mathbf{x} = \{\mathcal{I}, \mathcal{X}\}$ and ground truth output sequence $\mathbf{y}$, where $\mathcal{I} \in\mathbb{R}^{H\times W\times 3}$ represents a medical image and $\mathcal{X}=\{x_1,x_2,\dots,x_n, \mathbf{q}\}$ denotes $n$ serialized clinical records and the question $\mathbf{q}$, existing LVLMs first encode the visual and textual inputs separately, and then fuse them into a shared embedding space as
\begin{equation}
\mathbf{Z}^{(0)}=
\left[
f_{I\rightarrow t}(\mathcal{E}_I(\mathcal{I}))\,;\,\mathcal{E}_t(\mathcal{X})
\right]
\in\mathbb{R}^{(L_I+L_t)\times d_t}
\end{equation}
where $\mathcal{E}_I(\cdot)$ is a vision encoder and $\mathcal{E}_t(\cdot)$ is the textual encoder.
$L_I$ and $L_t$ denote the numbers of visual and textual tokens, respectively. $f_{I\rightarrow t}$ maps the visual embeddings into the shared embedding space, and $d_t$ denotes the dimensions of the shared embedding space.

The fused embedding is then processed by $L_b$ Transformer blocks $\mathcal{F}$,
$\mathbf{Z}^{(\ell)}=\mathcal{F}^{(\ell)}\!\left(\mathbf{Z}^{(\ell-1)}\right)$, $l=1,\dots,L_b$
yielding the final contextualized representation $\mathbf{Z}^{(L_b)}$. Conditioned on \(\mathbf{Z}^{(L_b)}\), the model generates the output sequence autoregressively with learnable parameters $\theta$ as
\begin{equation}
p_{\theta}(\mathbf{y}\mid \mathbf{x}) =
\prod_{i=1}^{|\mathbf{y}|}
p_{\theta}\!\left(y_i \mid y_{<i}, \mathbf{Z}^{(L_b)}\right)
\end{equation}

\subsection{Reinforcement Learning}
Reinforcement learning (RL) is a general framework designed for sequential decision-making, typically formulated as a Markov Decision Process (MDP) defined as
\begin{equation}
\mathcal{M} = (\mathcal{S}, \mathcal{A}, P, r, \gamma), \quad s \in \mathcal{S}, \; a \in \mathcal{A}
\end{equation}
where $\mathcal{S}$ denotes the state space, $\mathcal{A}$ denotes the action space, $P$ denotes the transition dynamics. $r$ is the reward function, and $\gamma \in [0,1]$ is the discount factor.
Given a state $s$, a policy $\pi_\phi(a \mid s)$ specifies a distribution over the action space.
The policy induces a trajectory $\tau$ with probability $p(\tau)$:
\begin{equation}
\tau = (s_1, a_1, s_2, a_2, \dots, s_T), \quad  p(\tau) = p(s_1)\prod_{k=1}^{T} \pi_\phi(a_k \mid s_k)\, P(s_{k+1} \mid s_k, a_k)
\end{equation}

The objective of RL is to learn a policy that maximizes the expected cumulative reward as
\begin{equation}
\max_{\pi_\phi} \; \mathbb{E}_{\tau \sim p_{\tau}(\cdot \mid \pi_\phi)}
\left[
\sum_{k=1}^{T} \gamma^{k-1} r(s_k, a_k)
\right],
\end{equation}
where $p_{\tau}(\cdot \mid \pi_\phi)$ denotes the trajectory distribution induced by policy $\pi_\phi$.


\section{Methodology}
\label{sec:methodology}

\subsection{\Method Overview}
\label{sec:overview}

\begin{figure}[t]
\begin{center}
\centerline{\includegraphics[width=\linewidth]{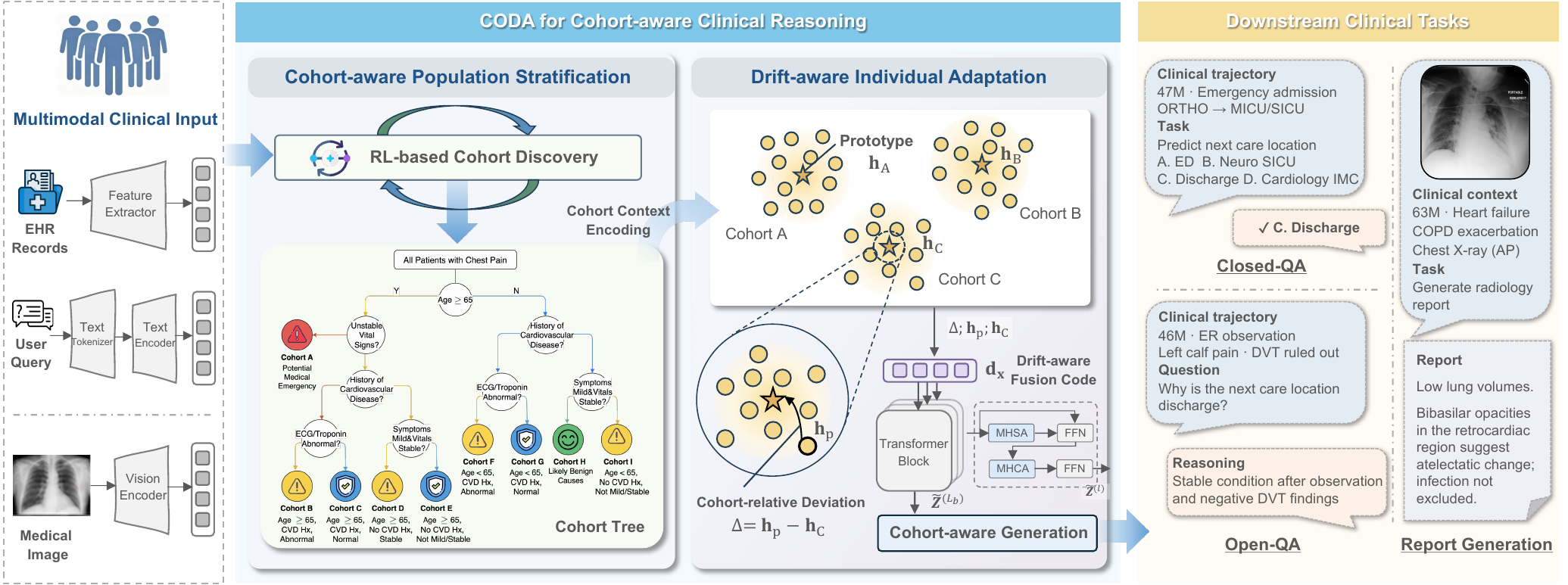}}
\caption{Overview of \Method for cohort- and drift-aware multimodal clinical reasoning.}
\label{fig:arch}
\end{center}
\vskip -0.3in
\end{figure}

As illustrated in Figure~\ref{fig:arch}, \Method adopts a two-stage formulation to enable cohort-aware clinical reasoning.
In the first stage, we learn a hierarchical cohort tree $\mathcal{T}$ that explicitly partitions the patient population $\mathcal{P}$ into a clinically meaningful cohort set $\mathcal{C}$. In the cohort tree, each internal node parameterizes a feature-based state $s$ and partition action $a$, while each leaf node represents a cohort $c \in \mathcal{C}$. These partition actions are optimized with reinforcement learning, allowing the cohort structure to be shaped not only by clinical heterogeneity but also by its utility for downstream generation. With the learned cohorts in $\mathcal{C}$, we derive a cohort-aware representation for each sample and inject it into the backbone, thereby providing explicit population-level context rather than relying on the model to infer such structure implicitly from its parameters alone.
In the second stage, building upon this cohort-level grounding, we maintain dynamically updated cohort prototypes $\mathbf{h}_c$ during training and characterize each sample by its cohort-relative deviation $\Delta(\mathbf{x},c)$. This cohort-relative deviation captures fine-grained patient-specific variability beyond shared cohort characteristics, enabling the model to reason jointly over stable population patterns and individualized clinical departures. The cohort- and drift-aware framework is encapsulated in the following integrated objective:
\begin{equation}
    \mathcal{J}(\theta, \pi_\phi) =\sum_{\mathbf{x}, \mathbf{y}}\Big[  \underbrace{\mathbb{E}_{\tau \sim p_{\tau}(\cdot \mid \pi_\phi)}[\mathcal{R}(\mathbf{x}, \tau)]+\mathcal{H}(p_{\pi_\phi}(\cdot | \mathbf{x}))}_{\text{Cohort-aware Population Stratification}} \quad+ \underbrace{\log p_\theta \big( \mathbf{y} \mid \mathbf{x}, \Delta(\mathbf{x} ; c) \big)}_{\text{Drift-aware Individual Adaptation}}\Big]
\end{equation}
where $\theta$ denotes the learnable model parameters and $\pi_\phi$ is the cohort-partitioning policy, $\mathcal{R}(\mathbf{x},\tau)$ is the trajectory-level cohort discovery reward,
$\mathcal{H}(p_{\pi_{\phi}}(\cdot \mid \mathbf{x}))$ is the policy entropy for cohort partition exploration.

\subsection{Cohort-aware Population Stratification}
To formalize the context-dependency inherent in clinical heuristics, \Method eschews the conventional assumption of a homogeneous patient distribution. Instead, \Method introduces a hierarchical stratification process that partitions the patient population space $\mathcal{P}$ into a set of clinically distinct cohorts $\mathcal{C}$. This process is governed by two synergistic components, i.e., an RL-based cohort discovery mechanism and a cohort context injection schema that injects cohort-level context into the model.

\noindent\textbf{RL-based Cohort Discovery.}
We formulate cohort discovery as a hierarchical policy learning problem, where a binary-tree-structured policy $\pi_{\phi}$ recursively partitions the patient population $\mathcal{P}$ into a set of cohorts $\mathcal{C}$.
Formally, we define a \emph{cohort tree} as a hierarchical binary tree 
$\mathcal{T} = (V, E)$, where $V=\{v_1,\cdots,v_N\}$ is the set of nodes and $E \subseteq V \times V$ is the set of directed edges representing parent–child relationships.
Each node $v \in V$ corresponds to a subset of patients $\mathcal{P}_v \subseteq \mathcal{P}$, with the root node representing the entire population.
Internal nodes correspond to decision points that further partition the patient subset, while leaf nodes represent terminal cohorts corresponding to clinically homogeneous patient groups.
Each node $v$ is further associated with (i) a state representation $s_v$ that summarizes the statistical and structural properties of the corresponding patient subset, and (ii) a partition action $a_v$ induced by the policy, which determines how the node is expanded or terminated.
The cohort tree thereby induces a recursive partition of the population space, where each root-to-leaf path encodes a sequence of feature-based decisions that uniquely determines a cohort.

Specifically, for each sample $\mathbf{x}$, we extract a clinical feature $\mathbf{u} = g(\mathbf{x})$, where $g(\cdot)$ maps serialized EHR records into a vector of statistical features capturing patient demographics, clinical conditions, and laboratory signals.
Each node $v$ is represented by a compact state vector:
\begin{equation}
    s_v =
    \left[
    \mu(\mathbf{U}_{\mathcal{P}_v});
    \sigma(\mathbf{U}_{\mathcal{P}_v});
    |\mathcal{P}_v|;
    d_v
    \right],
\end{equation}
where $\mathbf{U}_{\mathcal{P}_v}=\{\mathbf{u}_i\}_{i\in\mathcal{P}_v}$ denotes the collection of feature vectors of patients in node $v$. $\mu(\cdot)$ and $\sigma(\cdot)$ denote the mean and standard deviation, respectively.
$|\mathcal{P}_v|$ is the number of patients and $d_v$ is the depth of node $v$.

At each node $v$, the cohort policy $\pi_\phi$ induces a stochastic action $a_v$ over the patient subset $\mathcal{P}_v$ to specify whether to stop, the selected medical feature, and the corresponding split threshold, where action $a_v$ is defined as:
\begin{equation}
a_v = (z_v, j_v, \mathrm{t}_v) \sim \pi_{\phi}(\cdot \mid s_v).
\end{equation}
where $z_v \in \{0,1\}$ is a binary stopping variable, with $z_v=1$ indicating that $v$ becomes a terminal cohort. $j_v$ is the selected feature, and $t_v$ is the split threshold.
When $z_v = 0$, the node is expanded by partitioning $\mathcal{P}_v$ into two disjoint subsets:
\begin{equation}
\mathcal{P}_{v_L} = \{ i \in \mathcal{P}_v \mid u_{i,j_v} \le \mathrm{t}_v \}, 
\quad
\mathcal{P}_{v_R} = \mathcal{P}_v \setminus \mathcal{P}_{v_L},
\end{equation}
where $v_L$ and $v_R$ denote the left and right child nodes, respectively.


The reward function is designed to encourage each split to produce more compact and balanced child cohorts while penalizing unnecessary tree depth:
\begin{equation}
    r_v =
    \Big(
    \sigma(\mathbf{U}_{\mathcal{P}_v})
    -
    \sum_{k \in \{L,R\}}
    \frac{|\mathcal{P}_{v_k}|}{|\mathcal{P}_v|}
    \sigma(\mathbf{U}_{\mathcal{P}_{v_k}})
    \Big) +
    \left(
    1 - \frac{|\,|\mathcal{P}_{v_L}| - |\mathcal{P}_{v_R}|\,|}{|\mathcal{P}_v|}
    \right)
    -  \frac{d_v}{d_{\max}},
\end{equation}
where $d_{\max}$ is a normalization constant denoting the maximum depth. Here, the first term measures the reduction in intra-cohort variance after splitting, the second term discourages highly imbalanced partitions, and the last term penalizes overly deep trees. 

\noindent\textbf{Cohort Context Encoding.}
For each cohort $c$, we derive a representative embedding via a similarity-aware aggregation over its members, which is then mapped to a cohort embedding $\mathbf{e}_c$ and injected into the multimodal input sequence for the patients $\mathcal{P}_c$ within cohort $c$ to condition generation:
\begin{equation}
\mathbf{Z}^{(0)} =
\left[
f_{I \to t}(\mathcal{E}_I(\mathcal{I}));
\mathbf{e}_{c};
\mathcal{E}_t(\mathcal{X})
\right], \quad
\mathbf{e}_c = f_c\!\left(
\sum_{i \in \mathcal{P}_c}
\frac{\exp(\cos(\mathbf{h}_i,\bar{\mathbf{h}}_c))}
{\sum_{j \in \mathcal{P}_c}\exp(\cos(\mathbf{h}_j,\bar{\mathbf{h}}_c))}
\mathbf{h}_i
\right),
\end{equation}
where $f_c$ is a learnable mapping function,
$\mathbf{h}_i=f_{\mathbf{u}\rightarrow \mathbf{h}}(\mathbf{u}_i)$ is the learned patient representation based on statistical feature $\mathbf{u}_i$ via function $f_{\mathbf{u}\rightarrow \mathbf{h}}$, and $\bar{\mathbf{h}}_c$ is the centroid of cohort $c$ in the feature space.

To jointly optimize cohort discovery and cohort-aware generation, we define a unified objective that integrates policy learning and downstream generation:
\begin{equation}
\mathcal{L}_{\mathrm{cohort}}
=\mathbb{E}_{\tau \sim p_{\tau}(\cdot \mid \pi_\phi)}
\left[
\sum_{v\in V}
\left(
- \log \pi_\phi(a_v \mid \mathbf{s}_v) \, R_v
- \mathcal{H}(\pi_\phi(\cdot \mid \mathbf{s}_v))
\right)
\right]
+\lambda_g \mathcal{L}_{\mathrm{gen}},
\end{equation}
where $R_v = r_v - \mathcal{L}_{\mathrm{gen}}$ denotes the generation-aware reward and $\mathcal{H}(\pi_\phi(\cdot))$ is the entropy of the policy that encourages exploration.
$\mathcal{L}_{\mathrm{gen}}$ is the downstream generation loss with $\lambda_g$ balancing policy learning and generation performance.

In this way, a cohort tree $\mathcal{T}$ and cohort embeddings $\mathbf{e}_c$ are jointly learned. 
The resulting structure captures population-level heterogeneity via hierarchical partitioning, while the learned embeddings encode cohort-specific context for clinical reasoning. 
Through the proposed generation-aware objective, cohort discovery is directly aligned with multimodal reasoning, enabling the model to move beyond purely structure-driven partitioning toward reasoning-aware population stratification.

\subsection{Drift-aware Individual Adaptation}

While the cohort tree $\mathcal{T}$ provides population-level structure, clinical reasoning further requires capturing patient-specific deviations from cohort prototypes.
To this end, we introduce a drift-aware individual adaptation mechanism that explicitly models the cohort-relative deviation of each sample and integrates it into the generation process.

\noindent\textbf{Cohort Prototype Representation.}
For each cohort $c$, we maintain a learnable prototype representation $\mathbf{h}_c$, which summarizes the central clinical characteristics of the cohort and is initialized using the previously learned cohort embedding $\mathbf{e}_c$. 
During training, these prototypes are dynamically updated in an online manner based on the incoming samples. Specifically, for a sample $\mathbf{x}$ of patient $p$ assigned to cohort $c$ with feature representation $\mathbf{h}_p$, we update the corresponding prototype using an exponential moving average mechanism:
\begin{equation}
\mathbf{h}_{c} \leftarrow 
(1 - \beta \, w(p,c)) \mathbf{h}_{c}
+
\beta \, w(p,c) \, \frac{\mathbf{h}_p}{\|\mathbf{h}_p\|_2},
\end{equation}
where $w(p,c) \in [0,1]$ denotes the assignment confidence of patient $p$ to cohort $c$, and the feature is $\ell_2$-normalized to stabilize prototype updates. $\mathbf{h}_c$ replaces $\mathbf{e}_c$ in $\mathbf{Z}^{(0)}$ and serves as a conditioning signal in the multimodal input sequence, analogous to $\mathbf{e}_c$.

\noindent\textbf{Drift-aware Deviation Modeling.}
While the cohort tree $\mathcal{T}$ captures population-level structure, clinical reasoning further requires modeling patient-specific deviations from cohort prototypes. 
To this end, we introduce a unified drift-aware deviation mechanism that explicitly characterizes individual departures from cohort-level patterns and injects them into the generation process.

Given a sample $\mathbf{x}$ with feature representation $\mathbf{h}_p$ and assigned cohort $c$, we compute its cohort-relative deviation as $\Delta(\mathbf{x},c)=\mathbf{h}_p-\mathbf{h}_{c}$, which captures patient-specific variation beyond the cohort prototype. 
We first transform this deviation into a compact drift-aware fusion code:
\begin{equation}
\mathbf{d}_\mathbf{x}
=
\mathrm{LN}\!\left(
\mathbf{W}_d
\left[
\Delta(\mathbf{x},c);
\mathbf{h}_p;
\mathbf{h}_{c};
\Delta(\mathbf{x},c)\odot \mathbf{h}_{c}
\right]
+
\mathbf{b}_d
\right),
\end{equation}
where $[\cdot;\cdot]$ denotes concatenation, $\odot$ denotes element-wise interaction, $\mathrm{LN}(\cdot)$ denotes layer normalization, $\mathbf{W}_d$ is a learnable weight matrix and $\mathbf{b}_d$ is a learnable bias.

The drift code is then used to modulate the hidden representation at transformer block $\ell$ through a residual adapter. Compared to a simple additive bias, which applies a uniform shift, our formulation introduces a multiplicative interaction between the deviation and the hidden representation. This enables context-dependent, feature-wise modulation, allowing the model to adapt differently across features and tokens.
\begin{equation}
\tilde{\mathbf{Z}}^{(\ell)}
=
\mathbf{Z}^{(\ell)}
+
\mathbf{W}^{o}_{\ell}
\left(
\left(
1+\tanh(\mathbf{W}^{\gamma}_{\ell}\mathbf{d}_\mathbf{x})
\right)
\odot
\mathrm{GELU}(\mathbf{W}^{\alpha}_{\ell}\mathbf{Z}^{(\ell)})
+
\mathbf{W}^{\beta}_{\ell}\mathbf{d}_\mathbf{x}
\right),
\end{equation}
where $\mathbf{W}^{o}_{\ell}$ is a learnable output projection that maps the adapted signal back to the hidden space. 
The term $\mathrm{GELU}(\mathbf{W}^{\alpha}_{\ell}\mathbf{Z}^{(\ell)})$ corresponds to a nonlinear transformation of the transformer output, capturing content-dependent features. 
The modulation term $1+\tanh(\mathbf{W}^{\gamma}_{\ell}\mathbf{d}_\mathbf{x})$ acts as a feature-wise scaling factor conditioned on the cohort-relative deviation, where $\tanh(\cdot)$ ensures bounded and stable modulation. 
Finally, $\mathbf{W}^{\beta}_{\ell}\mathbf{d}_\mathbf{x}$ introduces an additive bias term that shifts the representation based on the deviation signal.

\noindent\textbf{Cohort-aware Generation.}
The adapted hidden representation $\tilde{\mathbf{Z}}^{(L_b)}$ at the final transformer block $L_b$ is then used for autoregressive generation:
\begin{equation}
p_{\theta}(\mathbf{y}\mid \mathbf{x}) =
\prod_{i=1}^{|\mathbf{y}|}
p_{\theta}\!\left(y_i \mid y_{<i}, \tilde{\mathbf{Z}}^{(L_b)}\right).
\end{equation}
To facilitate individual deviation modeling and enforce inter-cohort separability, we introduce a drift loss that explicitly encodes both intra-cohort consistency and inter-cohort discrimination by minimizing the deviation from the assigned cohort prototype while imposing a margin-based constraint against alternative cohorts.
\begin{equation}
    \mathcal{L}_{\mathrm{drift}}=\mathcal{L}_{\mathrm{gen}}+\lambda_{d}(\| \Delta(\mathbf{x},c)\|_2^2+\max(0, m-\frac{\mathbf{h}_p^\top \mathbf{h}_c}{\|\mathbf{h}_p\|\|\mathbf{h}_c\|}+\frac{\mathbf{h}_p^\top \mathbf{h}_{c'}}{\|\mathbf{h}_p\|\|\mathbf{h}_{c'}\|})) 
\end{equation}
where $m > 0$ is the cohort margin, $c'$ denotes the cohort with the second highest confidence to $p$, and $\lambda_d$ controls the trade-off between generation performance and cohort-relative deviation modeling.

Unlike standard generation, where the model relies solely on input features, our formulation conditions the generation process on both the cohort prototype $\mathbf{h}_{c}$ and the cohort-relative deviation $\Delta(\mathbf{x},c)$. 
This allows the model to first anchor predictions to population-level patterns, and then refine them based on patient-specific variations.
Overall, this design enables a unified integration of cohort-level regularities and individual-level deviations, leading to context-aware and fine-grained generation.
\begin{table}[t]
\centering
\tiny
\setlength{\tabcolsep}{4.2pt}
\renewcommand{\arraystretch}{0.9}
\caption{Comparison of \Method with baselines under \textit{Closed-QA} settings.}
\label{tab:exp-main-closed}
\resizebox{1\columnwidth}{!}{
\begin{tabular}{c lcccccccc cccc cc}
\toprule
\multirow{3}{*}{Domain} & \multirow{3}{*}{Model} & \multicolumn{8}{c}{ID} & \multicolumn{4}{c}{OOD} & \multicolumn{2}{c}{\multirow{2}{*}{Avg.}} \\
\cmidrule(lr){3-10}\cmidrule(lr){11-14}
&&\multicolumn{2}{c}{Admissions} &
\multicolumn{2}{c}{IE} &
\multicolumn{2}{c}{POE} &
\multicolumn{2}{c}{Services} &
\multicolumn{2}{c}{Transfers} &
\multicolumn{2}{c}{Obesity} &

\\
\cmidrule(lr){3-4}\cmidrule(lr){5-6}\cmidrule(lr){7-8}\cmidrule(lr){9-10}\cmidrule(lr){11-12}\cmidrule(lr){13-14}\cmidrule(lr){15-16}
& & ACC & MCC & ACC & MCC & ACC & MCC & ACC & MCC & ACC & MCC & ACC & MCC & ACC & MCC\\
\midrule
\multirow{8}{*}{\makecell[c]{General\\ Domain}} & LLaVA 1.5-7B&40.00 & 20.06 & 19.44 & -7.48 & 57.00 & 42.82 & 63.59 & 51.45 & {60.60} & {47.61} & 61.90 & 47.97 & 50.42 & 33.74\\
& LLaMA 3-11B& 32.80 & 10.74 & 26.52 & 4.85 & 35.40 & 14.55 & 43.64 & 25.32 & 37.00 & 16.29 & 54.12 & 39.58 & 38.25	& 18.56\\
& InternVL 3-8B& 28.20 & 5.06 & 18.94 & -6.52 & 46.80 & 29.13 & 42.39 & 23.35 & 27.00 & 2.58 & 70.92 & 59.61 & 39.04 & 18.87\\
& Qwen 3-VL-8B& 36.40 & 15.35 & 38.38 & 17.64 & 40.40 & 20.49 & 60.60 & 48.07 & 53.80 & 38.63 & 69.67 & 58.10 & 49.88 & 33.05\\
& Gemma 3-12B& 18.80 & -8.06 & 19.95 & -6.83 & 35.20 & 13.55 & 58.85 & 45.20 & 50.00 & 33.42 &{71.07}&60.24&42.31&22.92\\
& Grok 4.2& 23.60 & -1.58 & 31.57 & 8.56 & 35.20 & 13.43 & 25.19 & 0.17 & 40.60 & 20.45& 56.92&39.69&35.51&13.45 \\
& Claude 4.5 Haiku&35.60 & 14.09 & 29.29 & 6.51 & 40.40 & 20.78 & 42.64 & 23.54 & 42.60 & 23.51&47.90&32.61&39.74&20.17 \\
& GPT 5.4 Mini& 34.60 & 12.60 & 42.39 & 23.08 & 44.40 & 26.02 & 43.14 & 24.17 & 48.40 & 31.14 & 59.72 & 45.92 & 45.44 & 27.15\\
\midrule
\multirow{7}{*}{\makecell[c]{Medical\\ Domain}}& LLaVA-Med v1.5-7B & 21.00 & 3.83 & 25.00 & 2.21 & {58.00} & {45.01} & 44.89 & 27.44 & 34.40 & 17.29 & 58.48 & 45.85 & 40.29	& 23.60\\
& Med-Flamingo-7B& 17.40 & -3.86 & 8.08 & -3.11 & 21.80 & 3.18 & 25.69 & 4.28 & 27.80 & 1.84 & 17.42 & 3.60 & 19.70 & 0.99\\
& HuatuoGPT-Vision-7B& 44.00 & 26.00 & 40.66 & 20.78 & 48.40 & 32.11 & 60.10 & 46.94 & 57.80 & 44.60 & 67.96 & 56.04 &53.15&37.74\\
& HealthGPT-14B& 39.60 & 19.53 & 25.51 & 1.53 & 38.20 & 18.04 & 33.42 & 12.10 & 42.00 & 23.55 &55.68 & 42.71 & 39.07 & 19.58 \\
& Lingshu-7B& {44.40} & {26.26} & {45.45} & {28.06} & 50.60 & 34.10 & {78.55} & {71.56} & 57.00 & 44.26 & 70.63 & {60.52} & {57.77}	& {44.13}\\
& Hulu-Med-7B& 43.60 & 25.12 & 26.01 & 2.39 & 53.20 & 37.75 & 60.60 & 47.84 & 36.00 & 14.56 &65.47&53.56&47.48&30.20\\
\rowcolor{blue!10} \cellcolor{white}& \Method-Qwen3-8B &  \textbf{76.60} & \textbf{68.77} & \underline{92.42} & \underline{89.93} & \textbf{82.80} & \textbf{77.05} & \textbf{93.52} & \textbf{91.39} & \textbf{75.20} & \textbf{66.89} & \underline{72.32} & \underline{61.28} & \textbf{82.14} & \textbf{75.88}\\
\rowcolor{blue!10} \cellcolor{white}& \Method-Llama3-8B & \textbf{76.60} & \textbf{68.77} & \textbf{92.93} & \textbf{90.59} & 82.00 & 76.00 & \underline{92.27} & \underline{89.71} & \underline{74.80} & \underline{66.31} & \textbf{72.90} & \textbf{61.71} & \underline{81.92} & \underline{75.52}\\
\rowcolor{blue!10} \cellcolor{white}& \Method-Qwen3-8B w/o LoRA &  \underline{72.80} & \underline{63.75} & 92.17 & 89.59 & \underline{82.40} & \underline{76.53} & 85.29 & 80.55 & 70.20 & 60.61 & 72.16 & 60.98 & 79.17 & 72.00\\
\bottomrule
\end{tabular}
}
\vspace{-4mm}
\end{table}

\section{Experiments}
\label{sec:experiments}
\subsection{Experimental Setup}

\textbf{Datasets.}
We conduct extensive experiments on four real-world EHR benchmarks, i.e., EHR-Bench~\cite{liao2025ehr}, DIYHealth-Bench~\cite{liudiyhealth}, MIMIC-CXR~\cite{johnson2019mimic}, and PadChest~\cite{bustos2020padchest}.
Specifically, the admissions, services, ingredient events (IE), and provider order entry (POE) tasks from EHR-Bench, together with MIMIC-CXR, are treated as in-distribution (ID) datasets, while the Transfers task from EHR-Bench, Obesity from DIYHealth-Bench, and PadChest are used as out-of-distribution (OOD) datasets.
The model is trained on the ID training datasets and evaluated on both ID and OOD settings.
Through these datasets, we aim to provide a comprehensive assessment of the effectiveness and generalizability of \Method in answering practical medical questions and generating clinical reports.

\vspace{-1mm}
\textbf{Baselines.}
We evaluate \Method against a broad set of baselines, including state-of-the-art generalist models, (e.g., LLaVA 1.5~\citep{liu2023visual}, Llama 3~\citep{dubey2024llama}, InternVL 3~\citep{zhu2025internvl3}, Qwen3 VL~\citep{yang2025qwen3}, Gemma 3~\citep{team2025gemma}, Grok~\cite{xai2025grok4tech}, Claude 4.5 Haiku~\citep{anthropic2024claude} and GPT 5.4 Mini~\citep{achiam2023gpt}), as well as medical-specific models (e.g., LLaVA-Med v1.5~\citep{li2023llava}, Med-Flamingo~\citep{moor2023med}, HuatuoGPT-Vision~\citep{chen2024huatuogpt}, HealthGPT~\citep{lin2025healthgpt}, Lingshu~\citep{xu2025lingshu}, and Hulu-Med~\cite{jiang2025hulu}. All models are evaluated under closed-form multiple-choice question answering (Closed-QA), open-ended question reasoning (Open-QA), and radiology report generation. 

\vspace{-1mm}
\textbf{Training Strategy.}
Our training pipeline consists of three stages.
(i) \textit{Cross-Modal Alignment.} To adapt the visual and textual alignment to the medical domain, we train the vision projector of the backbone using a mixture of medical VQA and QA datasets, including MedQuAD~\cite{ben2019question}, PubMedVision~\cite{chen2024towards}, SLAKE~\cite{liu2021slake}, and VQA-RAD~\cite{lau2018dataset}.
(ii) \textit{Cohort-aware Population Stratification.} With the backbone frozen, we train only the RL-related modules on a subset (15\%) of the in-distribution data to learn a hierarchical cohort structure.
(iii) \textit{Drift-aware Individual Adaptation.} We fine-tune only the LoRA modules together with deviation-aware components to model patient-specific variations conditioned on cohort context.

\vspace{-1mm}
\textbf{Evaluation Metrics.}
For Closed-QA, we adopt Accuracy (ACC) and Matthews Correlation Coefficient (MCC)~\cite{matthews1975comparison}. ACC quantifies the proportion of correctly predicted answers, reflecting exact match performance. MCC provides a balanced evaluation by capturing the agreement between predicted and true labels across all classes. For Open-QA, we use BLEU-1~\citep{papineni2002bleu} and F1-BioBERT (F1-Bio)~\citep{zhangbertscore}. BLEU-1 evaluates lexical overlap based on unigram precision, serving as a signal of surface-level correctness. F1-Bio measures semantic consistency by computing similarity in a contextual embedding space derived from BioBERT~\citep{lee2020biobert}, enabling robust evaluation beyond surface-level matching.  For report generation, we employ F1-RadGraph (F1-Rad)~\citep{jain2021radgraph}  and F1-Bio. F1-Rad assesses the correctness of clinical entities and their relations by comparing structured representations extracted from RadGraph~\citep{jain2021radgraph}, capturing clinical fidelity at a fine-grained level. Please refer to Appendix~\ref{sec:appendix-implementation} for more details about the datasets, model implementation, and evaluation metrics.

\subsection{Main Results}
\vspace{-1mm}

\begin{table}[t]
\centering
\tiny
\setlength{\tabcolsep}{4.2pt}
\renewcommand{\arraystretch}{1}
\caption{Comparison of \Method with baselines under \textit{Open-QA} and \textit{report generation} settings.}
\label{tab:exp-main-open}
\resizebox{1\columnwidth}{!}{
\begin{tabular}{c lcccccccccc cccccc cccc}
\toprule
\multirow{3}{*}{Domain} & \multirow{3}{*}{Model} & \multicolumn{10}{c}{ID} & \multicolumn{6}{c}{OOD} & \multicolumn{4}{c}{\multirow{1}{*}{Avg.}} \\
\cmidrule(lr){3-12}\cmidrule(lr){13-18}\cmidrule(lr){19-22}
&&\multicolumn{2}{c}{Admissions} &
\multicolumn{2}{c}{IE} &
\multicolumn{2}{c}{POE} &
\multicolumn{2}{c}{Services} &
\multicolumn{2}{c}{MIMIC-CXR} &
\multicolumn{2}{c}{Transfers} &
\multicolumn{2}{c}{Obesity} &
\multicolumn{2}{c}{PadChest} &
\multicolumn{2}{c}{\multirow{1}{*}{Open-QA}} &
\multicolumn{2}{c}{\multirow{1}{*}{Report}}
\\
\cmidrule(lr){3-4}\cmidrule(lr){5-6}\cmidrule(lr){7-8}\cmidrule(lr){9-10}\cmidrule(lr){11-12}\cmidrule(lr){13-14}\cmidrule(lr){15-16}\cmidrule(lr){17-18}\cmidrule(lr){19-20}\cmidrule(lr){21-22}
& & F1-Bio & BLEU-1 & F1-Bio & BLEU-1 & F1-Bio & BLEU-1 & F1-Bio & BLEU-1 & F1-Rad & F1-Bio & F1-Bio & BLEU-1 & F1-Bio & BLEU-1& F1-Rad & F1-Bio& F1-Bio & BLEU-1 &F1-Rad & F1-Bio\\
\midrule
\multirow{8}{*}{\makecell[c]{General\\ Domain}} & LLaVA 1.5-7B&75.95 & 36.91 & 75.16 & 29.21 & 75.66 & 33.51 & 75.08 & 34.12 & 1.79 & 68.12 & 75.79 & 41.27 & 72.23 & 9.99 & 11.75 & 68.78 & 74.98 & 30.83 & 6.77 & 68.45\\
& LLaMA 3-11B& 74.68 & 35.65 & 72.73 & 24.38 & 74.83 & 36.18 & 74.97 & 34.34 & 4.38 & 65.01 & 74.88 & 38.01 & 70.51 & 8.68 & 7.54 & 62.28 & 73.77 & 29.54 & 5.96 & 63.64 \\
& InternVL 3-8B&72.20 & 24.43 & 73.91 & 25.01 & 72.61 & 23.90 & 72.39 & 23.83 & 6.38 & 61.62 & 73.96 & 26.16 & 67.90 & 5.34 & 9.32 & 61.45 & 72.16 & 21.45 & 7.85 & 61.54 \\
& Qwen 3-VL-8B& 68.82 & 18.31 & 68.77 & 17.14 & 68.47 & 16.82 & 68.07 & 17.09 & 8.23 & 63.63 & 72.23 & 26.96 & 68.95 & 6.40 & 8.15 & 66.42 & 69.22 & 17.12 & 8.19 & 65.03\\
& Gemma 3-12B& 72.33 & 26.56 & 71.86 & 25.30 & 73.57 & 30.07 & 74.57 & 33.98 & 10.27 & 68.75 & 74.72 & 25.58 & 66.93 & 4.07 & 12.31 & 65.35 & 72.33 & 24.26 & 11.29 & 67.05\\
& Grok 4.2& 69.61 & 19.71 & 68.97 & 19.71 & 68.61 & 17.82 & 69.65 & 19.86 & 13.52 & 72.62 & 71.21 & 25.32 & 67.63 & 5.29 & 12.74 & 70.31 & 69.28 & 17.95 & 13.13 & 71.46 \\
& Claude 4.5 Haiku&65.12 & 15.36 & 66.50 & 17.24 & 65.31 & 15.38 & 65.57 & 15.18 & 10.54 & 67.75 & 67.78 & 20.84 & 64.80 & 7.67 & 8.54 & 61.97 & 65.85 & 15.28 & 9.54 & 64.86 \\
& GPT 5.4 Mini& 68.86 & 23.15 & 65.55 & 26.80 & 69.59 & 23.77 & 68.84 & 22.84 & 15.88 & 72.30 & 71.24 & 30.65 & 65.78 & 7.07 & 10.35 & 68.39 & 68.31 & 22.38 & 13.11 & 70.34\\
\midrule
\multirow{7}{*}{\makecell[c]{Medical\\ Domain}}& LLaVA-Med v1.5-7B & 73.85 & 8.71 & 75.27 & 6.30 & 75.59 & 10.09 & 74.44 & 7.72 & 0.80 & 67.11 & 76.85 & 11.76 & 74.94 & 15.14 & 3.85 & 66.79 & 75.16 & 9.95 & 2.33 & 66.95\\
& Med-Flamingo-7B& 58.64 & 11.86 & 55.51 & 11.36 & 58.68 & 13.04 & 58.61 & 11.56 & 0.14 & 40.54 & 62.24 & 9.51 & 57.95 & 4.91 & 2.63 & 52.08 & 58.61 & 10.37 & 1.38 & 46.31\\
& HuatuoGPT-Vision-7B& 74.33 & 33.93 & 74.51 & 33.42 & 74.01 & 32.45 & 74.09 & 32.50 & 7.77 & 70.45 & 74.69 & 33.66 & 72.56 & 10.46 & 7.60 & 64.60 & 74.03 & 29.40 & 7.68 & 67.53 \\
& HealthGPT-14B& 69.91 & 29.27 & 70.67 & 27.16 & 71.03 & 26.87 & 72.15 & 28.22 & 9.05 & 73.04 & 66.58 & 27.69 & 66.49 & 7.49 & 2.03 & 68.48 & 69.47 & 24.45 & 5.54 & 70.76 \\
& Lingshu-7B& 76.17 & 38.17 & 74.68 & 35.01 & 75.88 & 33.47 & 76.26 & 33.23 & 16.18 & 78.19 & 76.11 & 35.19 & 72.95 & 11.16 & 1.23 & 68.85 & 75.34 & 31.04 & 8.71 & 73.52\\
& Hulu-Med-7B& 76.20 & 34.79 & 74.43 & 14.55 & 75.47 & 30.74 & 75.62 & 27.64 & 16.31 & 75.84 & 76.06 & 26.06 & 73.79 & 12.45 & 5.52 & 69.15 & 75.26 & 24.37 & 10.91 & 72.49 \\
\rowcolor{blue!10} \cellcolor{white}& \Method-Qwen3-8B &  \textbf{80.82} & \textbf{52.45} & \textbf{82.21} & \textbf{51.69} & \textbf{81.02} & \textbf{51.97} & \textbf{80.21} & \textbf{49.19} & \textbf{17.49} & \textbf{79.03} & \textbf{79.57} & \textbf{46.42} & \textbf{75.37} & \textbf{15.65} & \textbf{25.52} & \textbf{73.93} & \textbf{79.87} & \textbf{44.56} & \textbf{21.51} & \textbf{76.48}\\
\rowcolor{blue!10} \cellcolor{white}& \Method-Llama3-8B &78.49 & 46.38 & 78.74 & 46.11 & 77.30 & 46.05 & 76.35 & 44.51 & \underline{17.31} & \underline{78.56} & 77.49 & \underline{46.19} & \underline{74.95} & 14.52 & \underline{24.09} & \underline{71.77} & 77.22 & 40.63 & \underline{20.70} & \underline{75.17}\\
\rowcolor{blue!10} \cellcolor{white}& \Method-Qwen3-8B w/o LoRA & \underline{80.29} & \underline{50.50} &\underline{81.06} & \underline{48.43} & \underline{80.47} & \underline{51.64} & \underline{79.54} & \underline{48.37} & 16.76 & 78.08 & \underline{78.58} & 44.45 & 74.15 & \underline{14.82} & 23.32 & 71.64  & \underline{79.02} & \underline{43.04} & 20.04 & 74.86\\
\bottomrule
\end{tabular}
}
\vspace{-4mm}
\end{table}

The experimental results for Closed-QA, Open-QA, and report generation are presented in Tables~\ref{tab:exp-main-closed} and \ref{tab:exp-main-open}. We consider three variants: (i) \Method with a Qwen3 backbone, (ii) \Method with an LLaMA3 backbone, and (iii) \Method with a Qwen3 backbone but without LoRA. We summarize the key findings as follows. \textbf{(i) State-of-the-art performance.} Three variants of \Method consistently outperform all baselines across all tasks, yielding average gains of 42.18\% in ACC for Closed-QA, 6.01\% in F1-Bio for Open-QA, and 63.82\% in F1-Rad for report generation. These results demonstrate the effectiveness of \Method and suggest that explicitly modeling cohort structure and cohort drift provides a principled mechanism for capturing both population-level patterns and individual-level variations, resulting in consistently superior performance. \textbf{(ii) Out-of-Distribution Generalization.} Despite being trained solely on ID data, \Method generalizes well to OOD settings, as evidenced by its strong performance on the Obesity dataset, which represents a home-care scenario with limited EHR availability, as well as on PadChest, where the scale of EHR data is substantially smaller. These results indicate that \Method is robust to both data scarcity and domain shift, maintaining reliable performance across diverse clinical environments. \textbf{(iii) Robustness across backbones.} \Method consistently improves performance across diverse backbone architectures (e.g., Qwen3-VL and LLaMA-based models), demonstrating that the proposed cohort-aware mechanism is model-agnostic and broadly applicable. \textbf{(iv) Limitations of existing Med-LVLMs.} Existing Med-LVLMs are typically trained on large-scale corpora and a diverse set of tasks, largely mirroring general LVLM training paradigms with domain-specific data substitution. However, such models often overlook clinically meaningful reasoning processes and fail to incorporate domain-specific inductive biases or mechanisms. As a result, when data scale is no longer the primary bottleneck, their performance remains constrained by a lack of clinically grounded modeling.

\vspace{-2mm}
\subsection{Ablation Study and In-Depth Analysis}
\vspace{-2mm}

\begin{table}[t]
\centering
\tiny
\setlength{\tabcolsep}{4.2pt}
\renewcommand{\arraystretch}{1}
\caption{Ablation study results of \Method. "w/o Cohort" removes the Cohort Context Encoding, while "w/o Deviation" removes Drift-aware Deviation Modeling.}
\label{tab:ablation}
\resizebox{0.9\columnwidth}{!}{
\begin{tabular}{lcccccccc cccc}
\toprule
\multirow{3}{*}{Model} & \multicolumn{6}{c}{ID} & \multicolumn{6}{c}{OOD} \\
\cmidrule(lr){2-7}\cmidrule(lr){8-13}
&
\multicolumn{2}{c}{Closed-QA} &
\multicolumn{2}{c}{Open-QA} &
\multicolumn{2}{c}{Report} &
\multicolumn{2}{c}{Closed-QA} &
\multicolumn{2}{c}{Open-QA} &
\multicolumn{2}{c}{Report}
\\
\cmidrule(lr){2-3}\cmidrule(lr){4-5}\cmidrule(lr){6-7}\cmidrule(lr){8-9}\cmidrule(lr){10-11}\cmidrule(lr){12-13}
& ACC & MCC & F1-Bio & BLEU-1 & F1-Rad & F1-Bio & ACC & MCC & F1-Bio & BLEU-1 & F1-Rad & F1-Bio \\
\midrule
				
Vanilla Backbone &43.95 & 25.39 & 68.53 & 17.34 & 8.23 &63.63 & 61.74&48.36 & 70.59 & 16.68 & 8.15 &66.42\\
w/o Cohort& 85.58 & 80.69 & 80.11 & 49.18 & 15.73 & 78.58 & 72.56 & 62.19 & 76.82 & 28.83 & 24.60 & 72.51\\
w/o Deviation &85.56 & 80.73 & 80.10 & 48.86 & 16.33 & 78.94 & 73.05 & 62.95 & 76.65 & 28.82 & 25.19 & 72.41\\
\rowcolor{blue!10} \Method &\textbf{86.34} & \textbf{81.78} & \textbf{81.06} & \textbf{51.32} & \textbf{17.49} & \textbf{79.03} & \textbf{73.76} & \textbf{64.08} & \textbf{77.47} & \textbf{31.03} & \textbf{25.52} & \textbf{73.93}  \\
\bottomrule
\end{tabular}
}
\vspace{-3mm}
\end{table}

\noindent\textbf{Component-wise Ablation Study.}
To assess the contributions of key components in \Method, we conduct an ablation study, with results reported in Table~\ref{tab:ablation} and full results in Appendix~\ref{sec:appendix-ablation}. Removing the Cohort Context Encoding leads to consistent performance degradation in both in-distribution and out-of-distribution settings, particularly for Closed-QA and report generation tasks. This suggests that explicitly modeling cohort structure is essential for capturing population-level regularities. Removing deviation modeling results in additional performance drops, especially on Open-QA tasks, indicating that cohort-level representations alone are insufficient to capture fine-grained, patient-specific variations required for accurate clinical reasoning.
Overall, the full model (\Method) achieves the best performance across all settings, demonstrating that the joint modeling of cohort structure and individual-level deviations is critical for robust generalization.


\begin{figure}[t]
    \centering

    \begin{subfigure}[t]{0.32\textwidth}
        \centering
        \includegraphics[width=\linewidth]{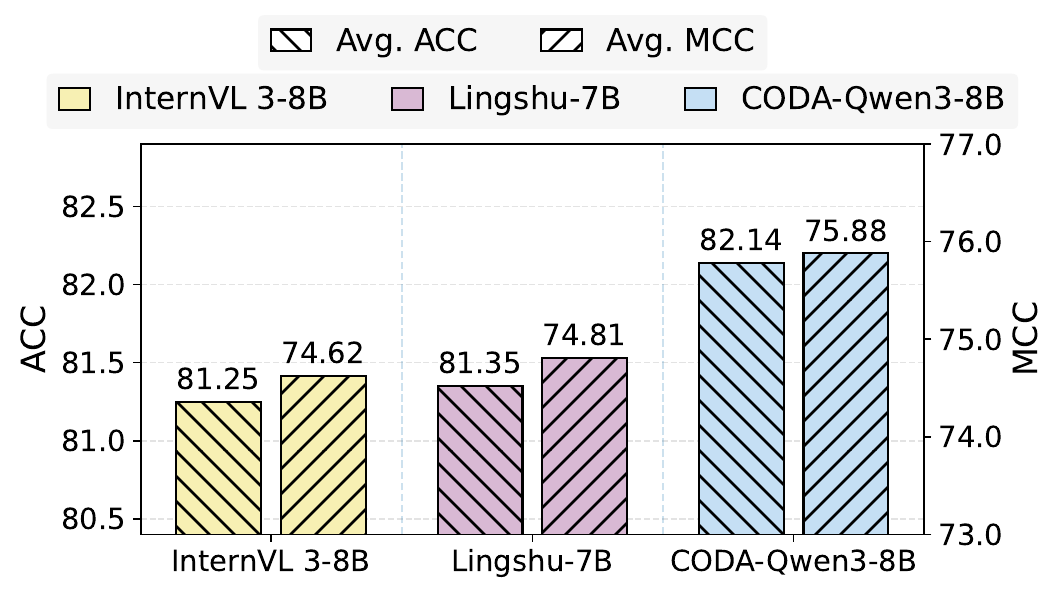}
        \vspace{-6mm}
        \caption{Results of \textit{Closed-QA}.}
        \label{fig:ablation-closed}
    \end{subfigure}
    \hfill
    \begin{subfigure}[t]{0.32\textwidth}
        \centering
        \includegraphics[width=\linewidth]{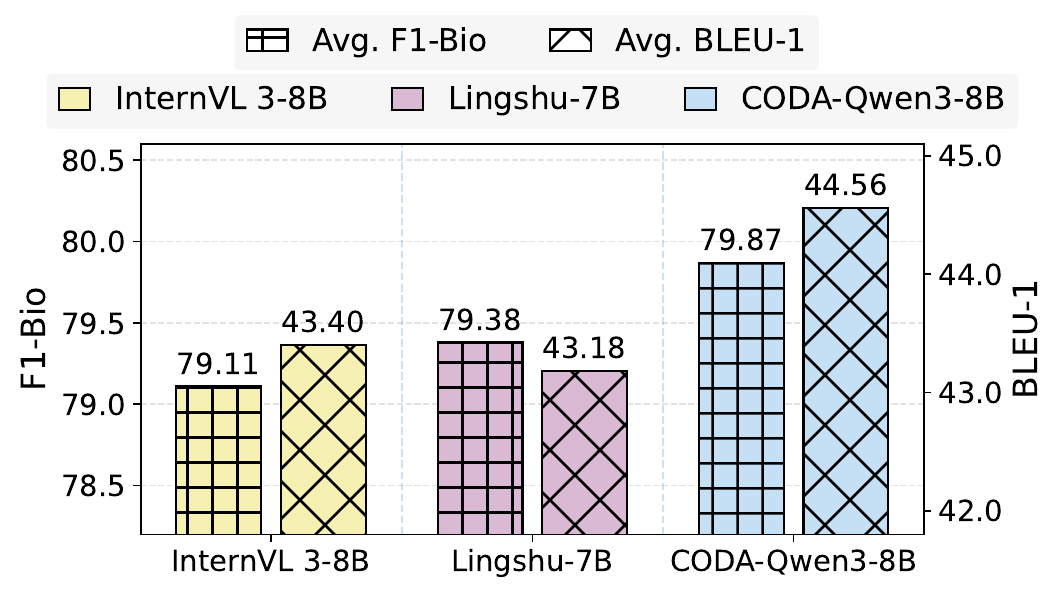}
        \vspace{-6mm}
        \caption{Results of \textit{Open-QA}.}
        \label{fig:ablation-open}
    \end{subfigure}
    \hfill
    \begin{subfigure}[t]{0.32\textwidth}
        \centering
        \includegraphics[width=\linewidth]{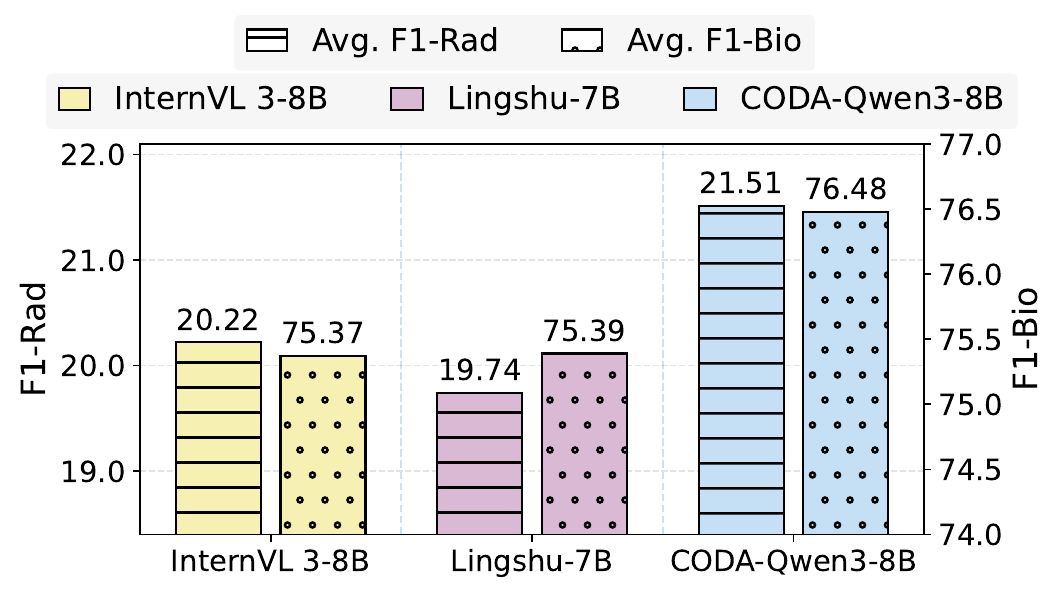}
        \vspace{-6mm}
        \caption{Results of \textit{Report Generation}.}
        \label{fig:ablation-report}
    \end{subfigure}

    \caption{Comparisons with fine-tuned baselines.}
    \label{fig:comparison_fine_tune}
    \vspace{-6mm}
\end{figure}

\begin{wrapfigure}[12]{t}{0.35\textwidth}
    \begin{center}
    \vspace{-8mm}
\includegraphics[width=0.34\textwidth]{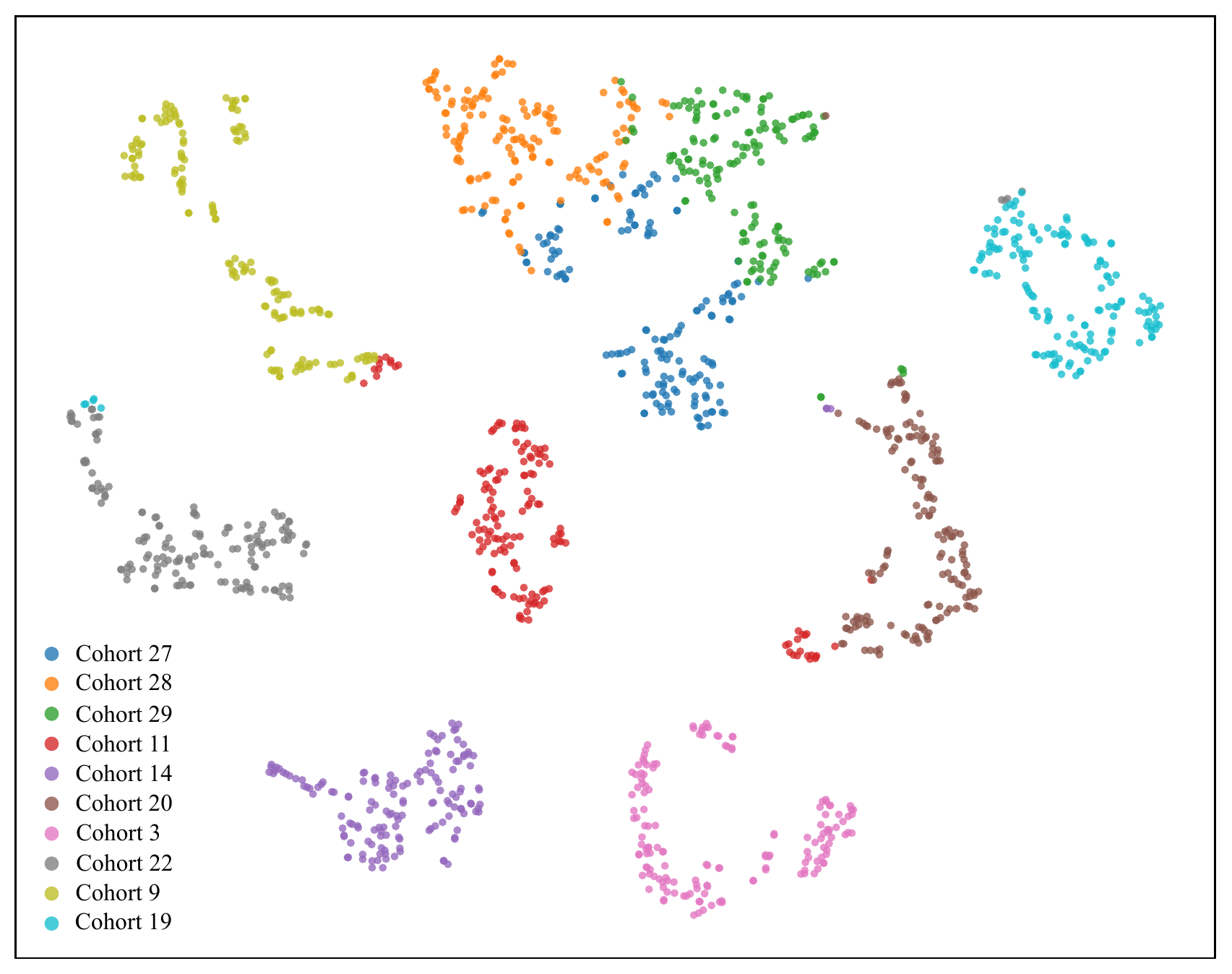}
        \vspace{-1mm}
        \caption{Visualization of embeddings within different cohorts.}
        \label{fig:tsne}
    \end{center}
\end{wrapfigure}
\noindent\textbf{Comparison with Fine-tuned Baselines.}
To fully validate the effectiveness of \Method, we compare \Method-Qwen3-8B with two representative baselines, Lingshu-7B and InternVL 3-8B, with results presented in Figure~\ref{fig:comparison_fine_tune} and full results in Appendix~\ref{sec:full-results-fine-tune}. \Method consistently achieves the best performance across all tasks and metrics. These results demonstrate that \Method not only improves predictive accuracy but also enhances semantic consistency and clinical fidelity, highlighting the effectiveness of its cohort-aware and deviation-aware modeling in complex medical reasoning tasks.
\noindent\textbf{Interpretation Study.}
We visualize the learned patient representations of \Method on OOD tasks using t-SNE~\cite{maaten2008visualizing}. For clarity, we present results from the top-10 largest cohorts. As shown in Figure~\ref{fig:tsne}, samples from the same cohort form compact clusters, while different cohorts are largely well separated, indicating intra-cohort consistency and inter-cohort separability. We also observe mild overlap among cohorts 27, 28, and 29, which can be attributed to their shared ancestor nodes in the cohort hierarchy, leading to partially aligned representations.
Overall, the learned embeddings exhibit coherent and discriminative geometric structures, where samples are grouped according to shared population-level patterns, suggesting that cohort semantics are effectively captured in the embedding space.




\noindent\textbf{Efficiency Study.}
To evaluate the computational efficiency of \Method, we compare it with representative general and medical baselines.
The results are summarized in Table~\ref{tab:efficiency_comparison}.
Despite introducing cohort-aware population stratification and drift-aware individual adaptation, \Method maintains competitive inference efficiency while achieving substantially stronger clinical reasoning performance.
Specifically, \Method achieves 7.95 s/sample and 20.19 tokens/s throughput, remaining comparable to widely used LLaVA-based baselines while exhibiting substantially lower sample latency than models with higher output complexity, such as InternVL3-8B and Qwen3-VL-8B.
In addition, \Method maintains a practical memory footprint for standard single-GPU deployment, while the additional FLOPs/token introduced by cohort-aware conditioning remain modest.
Overall, the results demonstrate that \Method achieves a favorable effectiveness–efficiency trade-off, enabling enhanced population-grounded clinical reasoning without sacrificing practical inference efficiency.


\begin{table}[t]
\centering
\small
\caption{Efficiency comparison of \Method against different baselines.}
\label{tab:efficiency_comparison}
\resizebox{0.9\linewidth}{!}{
\begin{tabular}{l ccccc}
\toprule
Model & Tokens & Latency & Token Throughput & Peak GPU Mem. & FLOPs \\
 & (tokens/sample) & (s/sample) & (tokens/s) & (GB) & (G/token) \\
\midrule
LLaVA 1.5-7B     &187.31 & 8.62 & 21.73 & 13.27 & 13.60 \\
LLaVA-Med v1.5-7B  &69.56   & 3.21 & 21.67 & 14.21 & 14.60 \\
Hulu-Med-7B     &72.75& 5.79 & 28.95 & 15.65 & 14.40 \\
Lingshu-7B       &167.62& 5.79 & 25.18 & 15.65 & 14.40 \\
InternVL 3-8B    &417.55& 19.53 & 21.38 & 15.00 & 14.80 \\
Qwen 3-VL-8B     &545.65& 23.15 & 23.57 & 16.47 & 15.03  \\
\rowcolor{blue!10} \Method-8B &160.51 & 7.95 & 20.19 & 17.25 & 15.69\\
\bottomrule
\end{tabular}
}
\vspace{-4mm}
\end{table}

\subsection{Expert Review}
\label{sec:exps-expert}
\vspace{-1mm}
To assess the practical significance of \Method, we conduct an expert review involving both GPT-5.4 and human experts. Specifically, we randomly sample 250 Open-QA instances from the OOD Transfers task and 250 report generation instances from the OOD PadChest task. We compare against five strong baselines, including Grok4.2, Hulu-Med, Qwen3 VL, Lingshu, and LLaVA 1.5.

\begin{wrapfigure}[10]{t}{0.6\textwidth}
   \centering
    \begin{subfigure}[t]{0.28\textwidth}
        \centering
    \includegraphics[width=\linewidth]{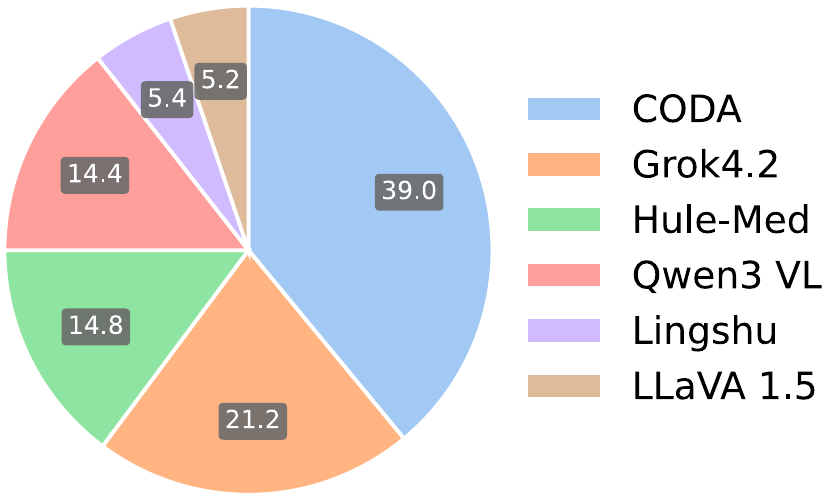}
    \vspace{-5mm}
        \caption{\small Results of human review on the first preference.}
        \label{fig:exper-review}
    \end{subfigure}
    \hfill
    \begin{subfigure}[t]{0.28\textwidth}
        \centering
    \includegraphics[width=\linewidth]{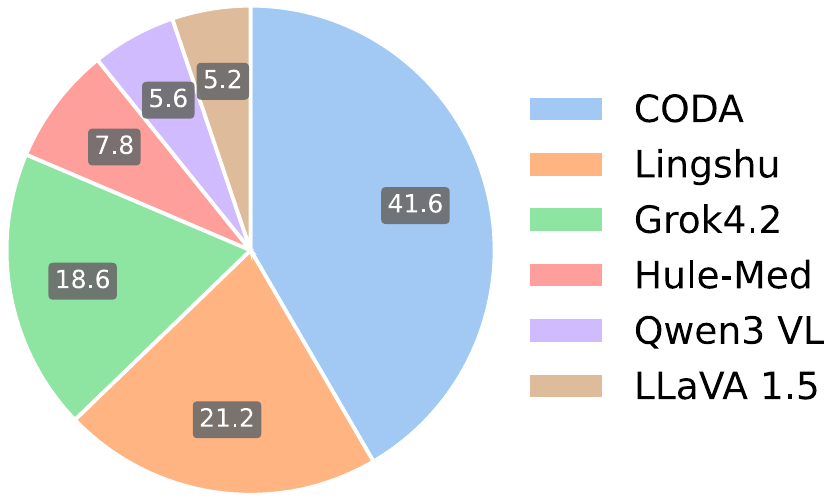}
    \vspace{-5mm}
        \caption{\small Results of GPT-5.4 review on the first preference.}
        \label{fig:gpt-review}
    \end{subfigure}
    \vspace{-1mm}
    \caption{\small Expert review results on the first preference.
    }
    \label{fig:review}
    \vspace{-4mm}
\end{wrapfigure}
For each sample, experts are asked to select their first preference, i.e., the response that is most accurate and clinically reliable, from six candidate answers. The results are presented in Figure~\ref{fig:review}. \Method is consistently chosen as the first preference by both human experts and GPT-5.4, demonstrating its superior clinical utility and robustness. Among the baselines, Grok-4.2 and Lingshu are ranked as the second-best models by human experts and GPT-5.4, respectively. Despite these variations, all baselines remain substantially behind \Method, indicating that the proposed cohort- and drift-aware mechanisms provide consistent and reliable improvements.

We provide additional in-depth analyses on few-shot settings, hyperparameter sensitivity, and case studies in Appendices~\ref{sec:appendix-few}, \ref{sec:appendix-hyperparameter}, and~\ref{sec:appendix-case}, further supporting the generality and reliability of \Method.
\section{Conclusions}
\label{sec:conclusion}
Generative AI has significantly advanced AI for healthcare.
However, current medical foundation models largely overlook the interplay between population-level cohort patterns and patient-specific variation that underlies real-world clinical reasoning.
In this paper, we propose \Method, a cohort- and drift-aware medical foundation model that explicitly discovers patient cohorts and integrates them into the reasoning pipeline.
\Method introduces cohort-aware population stratification and drift-aware individual adaptation, providing a principled and effective framework for enhancing medical reasoning.
Extensive experiments demonstrate that \Method consistently outperforms both general-purpose and medical-specific baselines across diverse clinical tasks, including closed-QA, open-QA, and report generation.
Despite its effectiveness, \Method currently leverages structured feature representations and summary statistics for cohort construction, which may introduce additional preprocessing overhead in certain scenarios. Developing more automated and end-to-end approaches for cohort discovery and integration remains an important direction for future work.

\small
\bibliographystyle{unsrt} 
\bibliography{reference}


\appendix
\newpage

\section*{Appendix}
\label{sec:appendix}


\begin{itemize}
  \item \textbf{\ref{sec:appendix-notation}. Notations} \dotfill \pageref{sec:appendix-notation}

  \item \textbf{\ref{sec:appendix-related-work}. Related Work} \dotfill \pageref{sec:appendix-related-work}

  \item \textbf{\ref{sec:appendix-implementation}. Implementation Details} \dotfill \pageref{sec:appendix-implementation}
  \begin{itemize}
      \item[$\cdot$] \textit{\ref{sec:appendix-dataset}. Datasets} \dotfill \pageref{sec:appendix-dataset}
      \item[$\cdot$] \textit{\ref{sec:appendix-model-implementation}. Model Implementation Details} \dotfill \pageref{sec:appendix-model-implementation}
      \item[$\cdot$] \textit{\ref{sec:appendix-environment}. Experimental Environment} \dotfill \pageref{sec:appendix-environment}
      \item[$\cdot$] \textit{\ref{sec:appendix-metrics}. Evaluation Metrics} \dotfill \pageref{sec:appendix-metrics}
  \end{itemize}

  \item \textbf{\ref{sec:appendix-cohort}. Cohort Structure} \dotfill \pageref{sec:appendix-cohort}
  
  \item \textbf{\ref{sec:full-results}. Supplementary Experimental Results} \dotfill \pageref{sec:full-results}
  \begin{itemize}
      \item[$\cdot$] \textit{\ref{sec:appendix-ablation}. Supplementary Ablation Study} \dotfill \pageref{sec:appendix-ablation}
      \item[$\cdot$] \textit{\ref{sec:full-results-fine-tune}. Supplementary Comparisons with Fine-tuned Baselines} \dotfill \pageref{sec:full-results-fine-tune}
      \item[$\cdot$] \textit{\ref{sec:appendix-few}. Comparison with Few-shot Baselines} \dotfill \pageref{sec:appendix-few}
      \item[$\cdot$] \textit{\ref{sec:appendix-hyperparameter}. Hyperparameter Study} \dotfill \pageref{sec:appendix-hyperparameter}
  \end{itemize}
  
\item \textbf{\ref{sec:appendix-case}. Case Study} \dotfill \pageref{sec:appendix-case}

\item \textbf{\ref{sec:appendix-impact}. Broader Impacts} \dotfill \pageref{sec:appendix-impact}
\end{itemize}


\newpage
\section{Notations}
\label{sec:appendix-notation}

To provide a comprehensive overview of the notations used throughout the paper, we present a summary of notations in Table~\ref{tab:notation} as a quick reference to facilitate the understanding and recall of each symbol.

\begin{table}[H]
   \small
    \centering
    \renewcommand{\arraystretch}{1.2}
    \caption{Notations.}
    \label{tab:notation}
    \resizebox{1\columnwidth}{!}{
        \begin{tabular}{c l}
    \toprule[1pt]
 Notation & Description\\
  \toprule[0.8pt]
$\mathbf{x}$ & Input instance, consisting of clinical records, a question and an image. \\
$\mathbf{y}$ & The ground truth output sequence.\\
$x_i$ & The i-th clinical records in $\mathbf{x}$.\\
$\mathbf{q}$ & The question in $\mathbf{x}$.\\
$\mathcal{I}$ & Input image of height H, width W, and three RGB channels.\\
$\mathcal{X}$ & The textual input, including clinical records and a question.\\
$\mathcal{Z}^{(0)}$ & Initial multimodal embedding, formed by visual and text embeddings.\\
$\mathcal{Z}^{(\ell)}$ & Output of $\ell$-th transformer block.\\
$L_b$& The number of transformer blocks in the backbone.\\
$f_{I\rightarrow t}$ & The function that maps the visual embeddings into the shared textual embedding space.\\
$\mathcal{E}_I(\cdot)$ & The visual encoder that maps images to embeddings.\\
$\mathcal{E}_t(\cdot)$ & The textual encoder that maps words to embeddings.\\
$L_I$ & The number of visual tokens in image $\mathcal{I}$, similar to $L_t$.\\
$d_t$& The dimensions of shared textual embedding space.\\
$p_\theta$ & The conditional distribution parameterized by a large language model with parameters $\theta$.\\
$\mathcal{M}$ & Markov decision process (MDP).\\
$\mathcal{S}$, $\mathcal{A}$, $P$, $r$ and $\gamma$& State space, action space, transition dynamics, reward function and discount factor, respectively.\\
$\pi_\phi$ & Policy, $\pi_\phi(a \mid s)$ parameterized by $\phi$, which in this work corresponds to the cohort-partitioning policy.\\
$\tau$ & Trajectory generated by the policy, i.e., a sequence of states and actions.\\
$\mathcal{F^{(\ell)}}$ & The $\ell$-th Transformer block.\\
$\mathcal{T}$& The learned hierarchical cohort tree.\\
$\mathcal{P}$ & The set of patients.\\
$\mathcal{C}$ & The set of cohorts.\\
$\mathbf{h}_\mathcal{C}$ & The cohort prototypes of cohorts.\\
$\Delta(\mathbf{x},c)$ & The deviation between input $\mathbf{x}$ and cohort $c$.\\
$\theta$ &  The learnable model parameters.\\
$p_{\tau}(\cdot \mid \pi_\phi)$ & The trajectory distribution induced by policy $\pi_\phi$\\
$\mathcal{R}(\mathbf{x},\tau)$ & The trajectory-level cohort discovery reward.\\
$\mathcal{H}(p)$ & Entropy of a distribution $p$. \\
$p_{\pi_{\phi}}(\cdot \mid \mathbf{x})$ & Policy distribution conditioned on input $\mathbf{x}$. \\
$g(\cdot)$ & The function that maps serialized EHR records into a vector of statistical features.\\
$V$ & The set of nodes in $\mathcal{T}$, similar to $E$. which is the set of edges in $\mathcal{T}$.\\
$\mu(\cdot)$ & The mean function.\\
$\sigma(\cdot)$& The standard deviation function.\\
$d_v$& Depth of node $v$.\\
$\mathbf{U}_{\mathcal{P}_v}$ & Set of statistical features of patients $\mathcal{P}_v$.\\
$z_v$&Binary stopping variable for node $v$.\\
$j_v$ & The selected feature for node $v$.\\
$\mathrm{t}_v$& The split threshold for node $v$.\\
$d_{\max}$ &The maximum depth.\\
$\mathbf{e}_c$& Cohort embedding that will be injected into the multimodal input. \\
$\mathcal{P}_c$& The patients within cohort $c$.\\
$\mathcal{L}_{\mathrm{gen}}$& Downstream generation loss.\\
$R_v$& The generation-aware reward.\\
$\mathbf{h}_p$& Learned representation of patient $p$.\\
$\mathbf{W}_{\ell}^\alpha$& Learnable weight matrix, similar to $\mathbf{W}_{\ell}^\beta$, $\mathbf{W}_{\ell}^\gamma$, $\mathbf{W}_{\ell}^o$ and $\mathbf{W}_d$.\\
    \bottomrule[1pt]
        \end{tabular}
        }
\end{table}

\newpage
\section{Related Work}
\label{sec:appendix-related-work}

\textbf{Foundation Models in General Domain.} The rapid development of LLMs has significantly reshaped modern AI, enabling strong general-purpose reasoning and coherent, human-like interactions. Beginning with generative pre-trained Transformers (GPT)~\cite{brown2020language, openai202334}, it has been widely proven that scaling model size and training data leads to substantial improvements in general-purpose question answering across a broad range of tasks~\cite{luo2022biogpt, luo2024biomedgpt, yang2023fingpt}. Subsequent advances further enhanced vision-language alignment and multimodal understanding through instruction tuning~\cite{wei2021finetuned, liu2023visual, wei2022chain} and reinforcement learning from human feedback (RLHF)~\cite{ouyang2022training}.
Recent studies explore multimodal alignment via Wasserstein barycenters~\cite{tang2026binding} and fine-grained image-text alignment for zero-shot anomaly detection~\cite{zhu2024llms}.
Beyond generative modeling, representation learning approaches, including embedding models such as E5~\cite{wang2022text} and dense retrieval frameworks like DPR~\cite{karpukhin2020dense}, have supported retrieval augmented generation (RAG)~\cite{lewis2020retrieval} by providing high-quality semantic representations. These approaches allow LLMs to incorporate external knowledge, thereby improving factual consistency and mitigating hallucinations. More recently, advanced LLMs, such as Claude~\cite{anthropic2024claude}, GPT-5~\cite{singh2025openai} and Gemini~\cite{team2023gemini}, have demonstrated strong performance in complicated multimodal understanding and generation tasks. Yet, the general-purpose LLMs are not readily applicable to specialized domains such as healthcare. In such high-stakes settings, effective reasoning requires not only general linguistic competence but also a deep understanding of domain-specific knowledge and complex clinical data.

\textbf{Foundation Models in Medical Domain.}
Following the success of general-purpose LLMs, extensive efforts have been made to adapt these models to the medical domain. Early efforts focus on domain-specific pretraining, where models such as BioBERT~\cite{lee2020biobert}, PubMedBERT~\cite{gu2021domain} and ClinicalBERT~\cite{huang2019clinicalbert} are trained on large-scale biomedical corpora to learn medical domain knowledge and preserve medical semantics. More recently, Med-PaLM~\cite{singhal2023large} and HuatuoGPT~\cite{zhang2023huatuogpt} are proposed to extend LLMs to clinical applications, including medical question answering, report generation, and medical decision support. At the same time, multimodal medical LVLMs are developed to integrate heterogeneous clinical data. For example, LLaVA-Med~\cite{li2023llava} and Med-Flamingo~\cite{moor2023med} enables joint reasoning over radiology images and medical reports. Building upon these advances, subsequent models, exemplified by HealthGPT~\cite{lin2025healthgpt} and Hulu-Med~\cite{jiang2025hulu}, further scale medical corpora and task diversity to enhance the medical understanding of complex clinical tasks involving both structured and unstructured data. Despite these improvements, existing models mainly operate at the individual instance level, treating each patient independently and relying on implicit medical knowledge learned by model parameters. However, real-world clinical reasoning is inherently cohort-based, where clinicians often compare patients within similar populations and reason about deviations from the typical patterns. The lack of explicit modeling of cohort structure and patient-level derivation limits the ability of current foundation models to capture clinically meaningful heterogeneity.

\section{Implementation Details}
\label{sec:appendix-implementation}

\subsection{Datasets}
\label{sec:appendix-dataset}
The experiments are conducted on four real-world EHR benchmarks: EHR-Bench~\cite{liao2025ehr}, DIYHealth-Bench~\cite{liudiyhealth}, MIMIC-CXR~\cite{johnson2019mimic}, and PadChest~\cite{bustos2020padchest}. The task description for each dataset is introduced in Table~\ref{tab:task_data}. EHR-Bench and DIYHealth-Bench provide raw QA pairs. For the MIMIC-CXR and PadChest datasets, we further preprocess the EHR records and radiology reports into unified QA-pair formats to ensure consistency with EHR-Bench and DIYHealth-Bench. To balance evaluation reliability and computational efficiency, datasets containing more than 2,500 samples are split by reserving 500 samples for testing and using the remaining samples for training. For smaller datasets, we adopt a 4:1 training-to-test split ratio.

\begin{table}[ht]
\centering
\renewcommand{\arraystretch}{1.4}
\caption{Task descriptions across datasets.}
\label{tab:task_data}
\resizebox{1\columnwidth}{!}{
\begin{tabular}{l|l}
\toprule
\multicolumn{1}{c|}{Dataset} & \multicolumn{1}{c}{Task Description} \\
\midrule
Admissions 
& Predict the next admission type from EHR records (Closed-QA) and provide clinical reasoning (Open-QA). \\

\midrule
Services 
& Predict the next clinical department from EHR records (Closed-QA) and provide clinical reasoning (Open-QA). \\ 

\midrule
Ingredient Events (IE) 
& Predict the next administered medication or substance from EHR records (Closed-QA) and provide clinical reasoning (Open-QA). \\

\midrule
Provider Order Entry (POE) 
& Predict the next clinical order type from EHR records (Closed-QA) and provide clinical reasoning (Open-QA). \\

\midrule
Transfers 
& Predict the next care location from EHR records (Closed-QA) and provide clinical reasoning (Open-QA). \\

\midrule
Obesity 
& Assess obesity risk from self-reported EHR records (Closed-QA) and provide supporting analysis (Open-QA). \\

\midrule
MIMIC-CXR
& Generate radiology reports from EHR records and chest X-ray images. \\

\midrule
PadChest 
& Generate radiology reports from a few EHR records and chest X-ray images. \\

\bottomrule
\end{tabular}
}
\end{table}

\newpage
\subsection{Model Implementation Details}
\label{sec:appendix-model-implementation}
We adopt Qwen3-8B and LLaMA3-8B as backbone models. 
The projector $f_{I\rightarrow t}$ is implemented as a two-layer MLP. 
Both $g(\cdot)$ and $f_c$ are instantiated as MLPs, and $f_{\mathbf{u}\rightarrow \mathbf{h}}$ is an MLP-based encoder. 
The maximum depth $d_{\max}$ is set to 100. We set $\lambda_g = 0.3$ and $\lambda_d=10^{-4}$ across all experiments.
The coefficient $\beta$ is set to 0.01 following prior work~\cite{grill2020bootstrap, tarvainen2017mean}, 
and the margin $m$ is set to 0.2 following prior study~\cite{schroff2015facenet}.  The assignment confidence of patient $p$ to cohort $c$ is computed based on the similarity between $\mathbf{h}_p$ and $\mathbf{h}_c$.

We present the detailed configurations for the training process. The specific settings are listed in Table~\ref{tab:train-implementation}. These hyperparameters are set up following prior studies~\cite{lin2025healthgpt, liu2023visual,li2023llava,liudiyhealth}. The architectural parameters are listed in Table~\ref{tab:architectural-parameters}. 

\begin{table}[ht]
\centering
\small
\renewcommand{\arraystretch}{1.2}
\caption{Summary of configurations for the training process.}
\resizebox{0.78\columnwidth}{!}{
\begin{tabular}{l|r|r|r}
\toprule
Configuration & Stage (i) & Stage (ii) & Stage (iii)  \\
\midrule
Optimizer & AdamW & AdamW & AdamW \\
Adapter LR & / & 5e-5 & 1e-4 \\
Learning Rate & 2e-5 & 1e-4 & 2e-5 \\
Weight Decay & / & 0.01 & 0.01 \\
Dropout Rate & 0.05 & 0.1 & 0.1 \\
LR Scheduler & Warmup & Constant & Cosine w/ Warmup \\
Max Sequence Length & 32k & 32k & 32k \\
\bottomrule
\end{tabular}
\label{tab:train-implementation}
}
\end{table}

\begin{table}[ht]
\centering
\small
\renewcommand{\arraystretch}{1}
\caption{Architectural parameters of \Method.}
\begin{tabular}{l|l|l}
\toprule
\textbf{Module} & \textbf{Parameter} & \textbf{Value / Shape} \\
\midrule

\textbf{Tokenizer}
 & Vocabulary size & 151{,}670 \\
\midrule

\multirow{5}{*}{\textbf{Backbone (Qwen3-VL)}}
 & Hidden size & 4096 \\
 & \# Transformer layers & 36 \\
 & Embedding matrix & (151670, 4096) \\
 & Self-attn projections 
 & q: (4096, 4096), k/v: (4096, 1024), o: (4096, 4096) \\
 & MLP 
 & gate/up: (4096, 12288), down: (12288, 4096) \\
\midrule

\multirow{3}{*}{\textbf{LoRA}}
 & Rank $r$ & 16 \\
 & Matrix $\mathbf{A}$ 
 & q/k/v/o/gate/up: (16, 4096), down: (16, 12288) \\
 & \multirow{2}{*}{Matrix $\mathbf{B}$}  
 & q/o: (4096,16), k/v: (1024,16), \\
 & 
 & gate/up: (12288,16), down: (4096,16) \\
\midrule

\multirow{4}{*}{\textbf{Stage ii}}
 & Cohort embedding dim & 1024 \\
 & Prefix length & 8 \\
 & Prefix projector 
 & (4096, 4096) $\rightarrow$ (4096, $8 \times 4096$) \\
 & Cohort encoder hidden dim & 4096 \\
\midrule

\multirow{5}{*}{\textbf{Stage iii}}
 & Drift dim & 1024 \\
 & Adapter bottleneck dim & 256 \\
 & Adapter (down / up) 
 & (4096, 256), (256, 4096) \\
 & Drift generator 
 & (1024, 512) \\
 & Gating module 
 & (1024, 4096) \\
\midrule

\multirow{2}{*}{\textbf{Cohort Memory}}
 & \# Cohorts & 34 \\
 & Prototype matrix & (34, 1024) \\

\bottomrule
\end{tabular}
\label{tab:architectural-parameters}
\end{table}

\subsection{Experimental Environment}
\label{sec:appendix-environment}
All experiments are conducted on a server equipped with an AMD EPYC 9334 CPU (32 cores, 2.7 GHz), 128 GB RAM, and an NVIDIA H100 NVL GPU with CUDA 12.9. 
The system runs Ubuntu 24.04 with Linux kernel 6.8.0-79-generic.

\subsection{Evaluation Metrics}
\label{sec:appendix-metrics}
\subsubsection{Evaluation Metrics for Closed-QA}
\begin{itemize}[leftmargin=*]
\item \textbf{Accuracy.} Accuracy (ACC) evaluates the overall prediction performance by computing the fraction of correctly classified instances among all samples. It provides a coarse-grained measure of classification effectiveness, especially under balanced label distributions.
\begin{equation}
    ACC= \frac{TP + TN}{TP + TN + FP + FN}
\end{equation}
where TP, TN, FP, and FN are the numbers of true positives, true negatives, false positives, and false negatives, respectively.
\item \textbf{Matthews Correlation Coefficient.} The Matthews Correlation Coefficient (MCC) provides a holistic evaluation of classification performance by capturing the correlation between predicted and true labels through the full confusion matrix. In contrast to accuracy, MCC remains robust under class imbalance, as it simultaneously considers all types of prediction outcomes and their interactions across classes. As a normalized correlation measure, MCC ranges from $-1$ to $1$, where $1$ indicates perfect alignment, $0$ corresponds to chance-level prediction, and $-1$ reflects complete disagreement.
\begin{equation}
\text{MCC} =
\frac{
N \sum_{k} C_{kk} - \sum_{k} T_k P_k
}{
\sqrt{
\left( N^2 - \sum_{k} P_k^2 \right)
\left( N^2 - \sum_{k} T_k^2 \right)
}
}
\end{equation}
where $C \in \mathbb{R}^{K \times K}$ denotes the confusion matrix, and $C_{kl}$ represents the number of samples whose ground-truth class is $k$ and predicted class is $l$. 
$T_k = \sum_{l} C_{kl}$ and $P_k = \sum_{l} C_{lk}$ denote the total number of true and predicted instances for class $k$, respectively. 
$N = \sum_{k}\sum_{l} C_{kl}$ is the total number of samples, and $K$ is the number of classes.

\end{itemize}
\subsubsection{Evaluation Metrics for Open-QA}
\begin{itemize}[leftmargin=*] 
\item \textbf{F1-BioBERT.} F1-BioBERT evaluates semantic consistency between generated answers and ground-truth responses by leveraging contextual embeddings from BioBERT~\citep{lee2020biobert}. Instead of relying on exact lexical matching, it computes similarity at the representation level, enabling robust assessment of paraphrased or semantically equivalent expressions in medical text. 

Specifically, precision and recall are defined based on semantic matching between predicted and reference tokens (or spans) in the embedding space, and the final score is computed as their harmonic mean:
\begin{equation}
\text{F1-BioBERT} = \frac{2 \cdot \text{Precision} \cdot \text{Recall}}{\text{Precision} + \text{Recall}}.
\end{equation}
Here, matches are determined according to similarity scores derived from BioBERT representations, with higher scores indicating stronger semantic alignment.
\item \textbf{BLEU-1.} BLEU-1 evaluates lexical fidelity by measuring the modified unigram precision between the generated text and the reference. Compared with higher-order BLEU variants, BLEU-1 focuses on word-level overlap and is therefore suitable for assessing whether key lexical content is preserved.
\begin{equation}
\text{BLEU-1} = \text{BP} \cdot \exp\left(\log p_1\right),
\end{equation}
where $p_1$ is the modified unigram precision:
\begin{equation}
p_1 =
\frac{
\sum_{w \in \mathcal{V}} \min \left( \mathrm{Count}_{\mathrm{pred}}(w), \mathrm{Count}_{\mathrm{ref}}(w) \right)
}{
\sum_{w \in \mathcal{V}} \mathrm{Count}_{\mathrm{pred}}(w)
}.
\end{equation}
Here, $\mathcal{V}$ denotes the unigram vocabulary, while $\mathrm{Count}_{\mathrm{pred}}(w)$ and $\mathrm{Count}_{\mathrm{ref}}(w)$ denote the frequency of unigram $w$ in the generated and reference texts, respectively. $\text{BP}$ is the brevity penalty that discourages overly short generations.
\subsubsection{Evaluation Metrics for Report Generation}
\end{itemize}
\begin{itemize}[leftmargin=*] 
\item \textbf{F1-RadGraph.} F1-RadGraph evaluates the quality of generated radiology reports by measuring the overlap between predicted and reference clinical graphs constructed using RadGraph~\citep{jain2021radgraph}. Specifically, it assesses the correctness of extracted medical entities and their associated relations, thereby capturing both factual accuracy and clinical consistency at the structured level. The metric is computed as the harmonic mean of precision and recall over matched graph elements:
\begin{equation}
\text{F1-RadGraph} = \frac{2 \cdot \text{Precision} \cdot \text{Recall}}{\text{Precision} + \text{Recall}}
\end{equation}
where precision and recall are defined based on the matching between predicted and ground-truth RadGraph annotations, with true positives corresponding to correctly identified entities and relations, and false positives/negatives representing mismatched or missing clinical concepts.
\end{itemize}

\section{Cohort Structure}
\label{sec:appendix-cohort}

\begin{figure}[ht]
\centering
\vspace{2mm}
\includegraphics[width=1\textwidth]{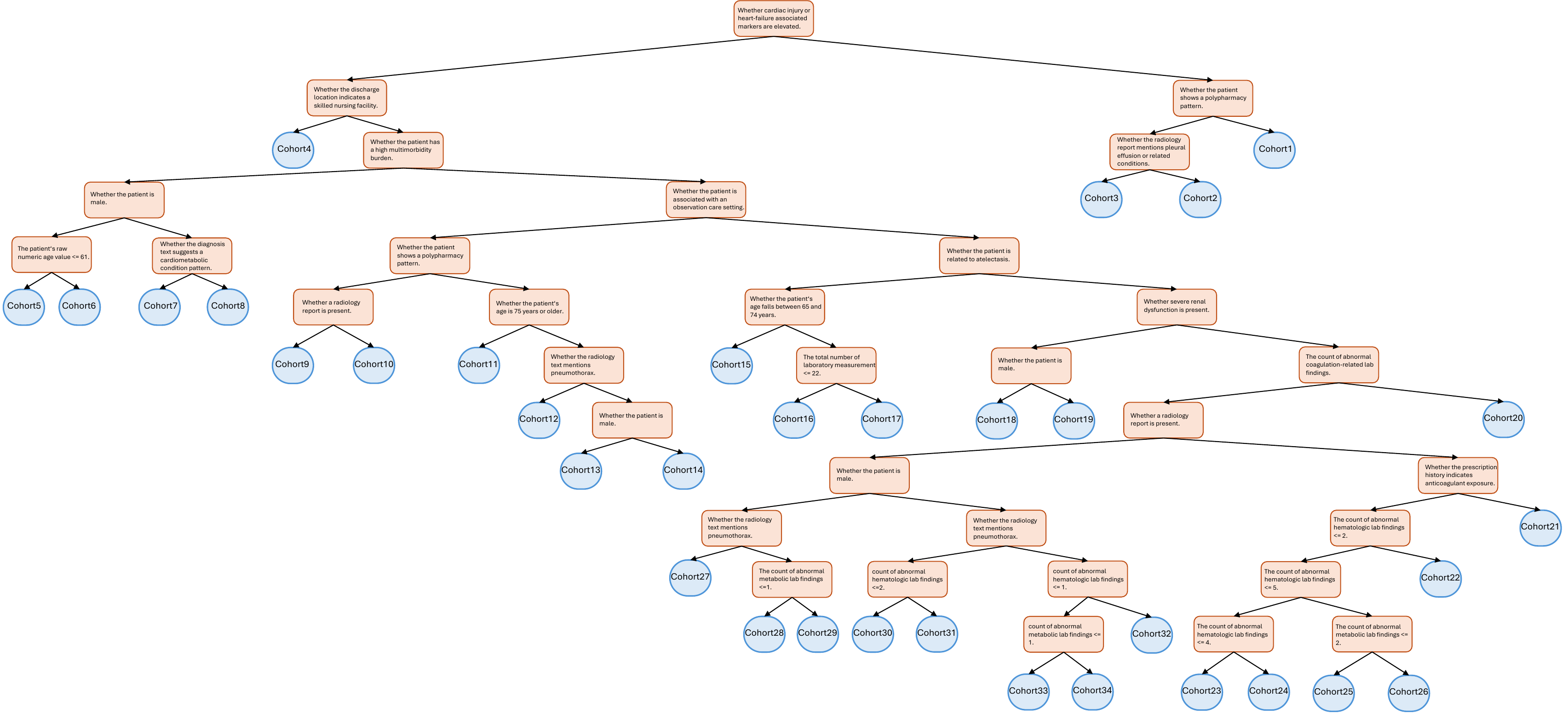}
\vspace{2mm}
\caption{Illustration of Cohort Structure.}
\label{fig:cohort-tree}
\end{figure}

The learned cohort tree provides an interpretable hierarchical decomposition of the patient space, where each internal node corresponds to a clinically meaningful decision criterion, and each leaf node defines a distinct cohort, as shown in Figure~\ref{fig:cohort-tree}. At the top levels, the model prioritizes global clinical signals such as cardiac-related markers, care context (e.g., discharge disposition and observation status), and multimorbidity, indicating that population stratification is first driven by overall severity and healthcare utilization. As the hierarchy progresses, the tree further branches along complementary axes, including chronic disease patterns (e.g., cardiometabolic conditions) and imaging-derived phenotypes (e.g., pleural effusion and pneumothorax), demonstrating that the model disentangles heterogeneous patient populations using semantically coherent, multimodal criteria. At deeper levels, the structure shifts toward fine-grained refinement, where demographic attributes (e.g., age and gender) and laboratory-derived signals (e.g., hematologic and metabolic abnormalities) are used to differentiate patients with subtle physiological variations. Notably, certain features recur across different branches, reflecting context-dependent decision boundaries rather than fixed global rules. In addition, the tree exhibits adaptive depth, with simpler cohorts identified early and more complex populations refined through deeper branches. 

Overall, this hierarchy reveals that the proposed method learns clinically grounded and interpretable cohort structures, organizing patients into meaningful subgroups that support coherent and reliable downstream reasoning.

\newpage
\section{Supplementary Experimental Results}
\label{sec:full-results}

\subsection{Supplementary Ablation Study}
\label{sec:appendix-ablation}

Tables~\ref{tab:ablation-closed} and~\ref{tab:ablation-open} provide extended ablation results across closed-QA, open-QA, and report generation settings.
Consistent with the main paper, removing either cohort-aware population stratification or drift-aware individual adaptation leads to performance degradation across both in-distribution and out-of-distribution tasks, demonstrating that the two components are complementary and jointly contribute to robust multimodal clinical reasoning.

Under closed-QA settings, CODA achieves the best overall performance across all tasks and evaluation metrics.
Similar trends are observed in open-QA and report generation settings. 
CODA consistently achieves the strongest performance across both language-level and clinical evaluation metrics.
Overall, the results demonstrate that population-level cohort structure and patient-specific deviation modeling are both critical for reliable multimodal clinical reasoning.

\begin{table}[ht]
\centering
\tiny
\setlength{\tabcolsep}{4.2pt}
\renewcommand{\arraystretch}{0.9}
\caption{Ablation Study of \Method under \textit{Closed-QA} settings.}
\label{tab:ablation-closed}
\resizebox{1\columnwidth}{!}{
\begin{tabular}{lcccccccc cccc cc}
\toprule
\multirow{3}{*}{Model} & \multicolumn{8}{c}{ID} & \multicolumn{4}{c}{OOD} & \multicolumn{2}{c}{\multirow{2}{*}{Avg.}} \\
\cmidrule(lr){2-9}\cmidrule(lr){10-13}
&\multicolumn{2}{c}{Admissions} &
\multicolumn{2}{c}{IE} &
\multicolumn{2}{c}{POE} &
\multicolumn{2}{c}{Services} &
\multicolumn{2}{c}{Transfers} &
\multicolumn{2}{c}{Obesity} &
\\
\cmidrule(lr){2-3}\cmidrule(lr){4-5}\cmidrule(lr){6-7}\cmidrule(lr){8-9}\cmidrule(lr){10-11}\cmidrule(lr){12-13}\cmidrule(lr){14-15}
& ACC & MCC & ACC & MCC & ACC & MCC & ACC & MCC & ACC & MCC & ACC & MCC & ACC & MCC\\
\midrule
LoRA Only&74.60 & 65.78 & 91.92 & 89.04 & 81.20 & 74.83 & 92.02 & 89.42 & 73.60 & 64.65 & 70.30 & 58.48 & 80.61 & 73.70\\
w/o Cohort&75.20 & 66.57 & 92.17 & 89.57 & 82.20 & 76.25 & 92.77 & 90.38 & 74.20 & 65.16 & 70.92 & 59.21 &  81.24 & 74.52\\
w/o Deviation & 75.80 & 67.37 & 92.17 & 89.51 & 82.00 & 76.00 & 92.27 & 90.06 & 74.40 & 65.70 & 71.70 & 60.20  & 81.39 & 74.81\\
\rowcolor{blue!10} \Method &  \textbf{76.60} & \textbf{68.77} & \textbf{92.42} & \textbf{89.93} & \textbf{82.80} & \textbf{77.05} & \textbf{93.52} & \textbf{91.39} & \textbf{75.20} & \textbf{66.89} & \textbf{72.32} & \textbf{61.28} & \textbf{82.14} & \textbf{75.88}\\
\bottomrule
\end{tabular}
}
\end{table}

\begin{table}[ht]
\centering
\tiny
\setlength{\tabcolsep}{4.2pt}
\renewcommand{\arraystretch}{0.9}
\caption{Ablation Study of \Method under \textit{Open-QA} and \textit{report generation} settings.}
\label{tab:ablation-open}
\resizebox{1\columnwidth}{!}{
\begin{tabular}{lcccccccccc cccccc cccc}
\toprule
\multirow{3}{*}{Model} & \multicolumn{10}{c}{ID} & \multicolumn{6}{c}{OOD} & \multicolumn{4}{c}{\multirow{1}{*}{Avg.}} \\
\cmidrule(lr){2-11}\cmidrule(lr){12-17}\cmidrule(lr){18-21}
&\multicolumn{2}{c}{Admissions} &
\multicolumn{2}{c}{IE} &
\multicolumn{2}{c}{POE} &
\multicolumn{2}{c}{Services} &
\multicolumn{2}{c}{MIMIC-CXR} &
\multicolumn{2}{c}{Transfers} &
\multicolumn{2}{c}{Obesity} &
\multicolumn{2}{c}{PadChest} &
\multicolumn{2}{c}{\multirow{1}{*}{Open-QA}} &
\multicolumn{2}{c}{\multirow{1}{*}{Report}}
\\
\cmidrule(lr){2-3}\cmidrule(lr){4-5}\cmidrule(lr){6-7}\cmidrule(lr){8-9}\cmidrule(lr){10-11}\cmidrule(lr){12-13}\cmidrule(lr){14-15}\cmidrule(lr){16-17}\cmidrule(lr){18-19}\cmidrule(lr){20-21}
& F1-Bio & BLEU-1 & F1-Bio & BLEU-1 & F1-Bio & BLEU-1 & F1-Bio & BLEU-1 & F1-Rad & F1-Bio & F1-Bio & BLEU-1 & F1-Bio & BLEU-1& F1-Rad & F1-Bio& F1-Bio & BLEU-1 &F1-Rad & F1-Bio\\
\midrule
LoRA Only&79.06 & 46.09 & 80.94 & 47.20 & 80.16 & 50.27 & 78.87 & 46.24 & 15.46 & 78.74 & 78.61 & 44.18 & 74.61 & 13.44 & 24.68 & 72.24 & 78.71 & 41.24 & 20.07 & 75.49\\
 w/o Cohort&79.71 & 48.02 & 81.10 & 49.08 & 80.44 & 52.03 & 79.20 & 47.60 & 15.73 & 78.58 & 78.58 & 43.66 & 75.07 & 14.01 & 24.60 & 72.51 & 79.02 & 42.40 & 20.17 & 75.55\\
w/o Deviation & 79.80 & 48.64 & 81.18 & 48.99 & 80.06 & 50.50 & 79.34 & 47.32 & 16.33 & 78.94 & 78.55 & 43.68 & 74.76 & 13.96 & 25.19 & 72.41 & 78.95 & 42.18 & 20.76 & 75.67\\
\rowcolor{blue!10} \Method & \textbf{80.82} & \textbf{52.45} & \textbf{82.21} & \textbf{51.69} & \textbf{81.02} & \textbf{51.97} & \textbf{80.21} & \textbf{49.19} & \textbf{17.49} & \textbf{79.03} & \textbf{79.57} & \textbf{46.42} & \textbf{75.37} & \textbf{15.65} & \textbf{25.52} & \textbf{73.93}  & \textbf{79.87} & \textbf{44.56} & \textbf{21.51} & \textbf{76.48}\\
\bottomrule
\end{tabular}
}
\end{table}

\subsection{Supplementary Comparisons with Fine-tuned Baselines}
\label{sec:full-results-fine-tune}

Tables~\ref{tab:fine-tuned-closed} and~\ref{tab:fine-tuned-open}  provide the full comparisons between CODA and representative LoRA fine-tuned multimodal medical foundation models under closed-QA, open-QA, and report generation settings.
CODA consistently achieves the best overall performance across both in-distribution and out-of-distribution tasks.

\begin{table}[H]
\centering
\tiny
\setlength{\tabcolsep}{4.2pt}
\renewcommand{\arraystretch}{0.9}
\caption{Comparison of \Method with LoRA fine-tuned baselines under \textit{Closed-QA} settings.}
\label{tab:fine-tuned-closed}
\resizebox{1\columnwidth}{!}{
\begin{tabular}{lcccccccc cccc cc}
\toprule
\multirow{3}{*}{Model} & \multicolumn{8}{c}{ID} & \multicolumn{4}{c}{OOD} & \multicolumn{2}{c}{\multirow{2}{*}{Avg.}} \\
\cmidrule(lr){2-9}\cmidrule(lr){10-13}
&\multicolumn{2}{c}{Admissions} &
\multicolumn{2}{c}{IE} &
\multicolumn{2}{c}{POE} &
\multicolumn{2}{c}{Services} &
\multicolumn{2}{c}{Transfers} &
\multicolumn{2}{c}{Obesity} &
\\
\cmidrule(lr){2-3}\cmidrule(lr){4-5}\cmidrule(lr){6-7}\cmidrule(lr){8-9}\cmidrule(lr){10-11}\cmidrule(lr){12-13}\cmidrule(lr){14-15}
& ACC & MCC & ACC & MCC & ACC & MCC & ACC & MCC & ACC & MCC & ACC & MCC & ACC & MCC\\
\midrule
InternVL 3-8B&76.00 & 67.98 & 91.18 & 88.51 & 81.20 & 74.93 & 93.27 & 91.02 & 74.60 & 65.75 & 71.23 & 59.52 & 81.25 & 74.62\\
Lingshu-7B & 76.20 & 68.24 & 91.67 & 88.85 & 81.20 & 74.96 & 93.27 & 91.07 & 74.40 & 65.52 & 71.38 & 60.24 & 81.35 & 74.81\\
\rowcolor{blue!10} \Method-Qwen3-8B &  \textbf{76.60} & \textbf{68.77} & \textbf{92.42} & \textbf{89.93} & \textbf{82.80} & \textbf{77.05} & \textbf{93.52} & \textbf{91.39} & \textbf{75.20} & \textbf{66.89} & \textbf{72.32} & \textbf{61.28} & \textbf{82.14} & \textbf{75.88}\\
\bottomrule
\end{tabular}
}
\end{table}

\begin{table}[H]
\centering
\tiny
\setlength{\tabcolsep}{4.2pt}
\renewcommand{\arraystretch}{0.9}
\caption{Comparison of \Method with LoRA fine-tuned baselines under \textit{Open-QA} and \textit{report generation} settings.}
\label{tab:fine-tuned-open}
\resizebox{1\columnwidth}{!}{
\begin{tabular}{lcccccccccc cccccc cccc}
\toprule
\multirow{3}{*}{Model} & \multicolumn{10}{c}{ID} & \multicolumn{6}{c}{OOD} & \multicolumn{4}{c}{\multirow{1}{*}{Avg.}} \\
\cmidrule(lr){2-11}\cmidrule(lr){12-17}\cmidrule(lr){18-21}
&\multicolumn{2}{c}{Admissions} &
\multicolumn{2}{c}{IE} &
\multicolumn{2}{c}{POE} &
\multicolumn{2}{c}{Services} &
\multicolumn{2}{c}{MIMIC-CXR} &
\multicolumn{2}{c}{Transfers} &
\multicolumn{2}{c}{Obesity} &
\multicolumn{2}{c}{PadChest} &
\multicolumn{2}{c}{\multirow{1}{*}{Open-QA}} &
\multicolumn{2}{c}{\multirow{1}{*}{Report}}
\\
\cmidrule(lr){2-3}\cmidrule(lr){4-5}\cmidrule(lr){6-7}\cmidrule(lr){8-9}\cmidrule(lr){10-11}\cmidrule(lr){12-13}\cmidrule(lr){14-15}\cmidrule(lr){16-17}\cmidrule(lr){18-19}\cmidrule(lr){20-21}
& F1-Bio & BLEU-1 & F1-Bio & BLEU-1 & F1-Bio & BLEU-1 & F1-Bio & BLEU-1 & F1-Rad & F1-Bio & F1-Bio & BLEU-1 & F1-Bio & BLEU-1& F1-Rad & F1-Bio& F1-Bio & BLEU-1 &F1-Rad & F1-Bio\\
\midrule
 InternVL 3-8B&80.73 & 52.10 & 81.54 & 50.22 & 80.91 & 51.51 & 80.18 & \textbf{49.39} & 16.59 & 78.69 & 79.14 & 46.00 & 72.17 & 11.21 & 23.85 & 72.05 & 79.11 & 43.74 & 20.22 & 75.37\\
Lingshu-7B & 80.60 & 48.92 & 82.16 & 50.76 & 80.99 & 51.65 & 80.01 & 49.03 & 16.89 & 78.88 & 78.82 & \textbf{46.64} & 73.72 & 12.07 & 22.58 & 71.90 & 79.38 & 43.44 & 19.74 & 75.39\\
\rowcolor{blue!10} \Method & \textbf{80.82} & \textbf{52.45} & \textbf{82.21} & \textbf{51.69} & \textbf{81.02} & \textbf{51.97} & \textbf{80.21} & 49.19 & \textbf{17.49} & \textbf{79.03} & \textbf{79.57} & 46.42 & \textbf{75.37} & \textbf{15.65} & \textbf{25.52} & \textbf{73.93}  & \textbf{79.87} & \textbf{44.56} & \textbf{21.51} & \textbf{76.48}\\
\bottomrule
\end{tabular}
}
\end{table}

\newpage
\subsection{Comparison with Few-shot Baselines}
\label{sec:appendix-few}
To further evaluate \Method against strong baselines, we augment several representative models with few-shot demonstrations. The results are reported in Tables~\ref{tab:few-shot-closed} and \ref{tab:few-shot-open}. Due to the rich and heterogeneous nature of EHR data, each sample is relatively long, posing challenges for models with limited context length constraints. While InternVL 3 and Hulu-Med benefit from few-shot demonstrations, LLaVA 1.5 and Lingshu exhibit performance degradation, suggesting insufficient capability to effectively process long-context inputs. In particular, the extreme degradation observed in LLaVA 1.5 under the 2-shot setting can also be attributed to its 4K context window, which is insufficient to accommodate both demonstrations and the query. Overall, these results further highlight the effectiveness of \Method.

\begin{table}[ht]
\centering
\tiny
\setlength{\tabcolsep}{4.2pt}
\renewcommand{\arraystretch}{0.9}
\caption{Comparison of \Method with few-shot baselines under \textit{Closed-QA} settings.}
\vspace{2mm}
\label{tab:few-shot-closed}
\resizebox{1\columnwidth}{!}{
\begin{tabular}{lcccccccc cccc cc}
\toprule
\multirow{3}{*}{Model} & \multicolumn{8}{c}{ID} & \multicolumn{4}{c}{OOD} & \multicolumn{2}{c}{\multirow{2}{*}{Avg.}} \\
\cmidrule(lr){2-9}\cmidrule(lr){10-13}
&\multicolumn{2}{c}{Admissions} &
\multicolumn{2}{c}{IE} &
\multicolumn{2}{c}{POE} &
\multicolumn{2}{c}{Services} &
\multicolumn{2}{c}{Transfers} &
\multicolumn{2}{c}{Obesity} &
\\
\cmidrule(lr){2-3}\cmidrule(lr){4-5}\cmidrule(lr){6-7}\cmidrule(lr){8-9}\cmidrule(lr){10-11}\cmidrule(lr){12-13}\cmidrule(lr){14-15}
& ACC & MCC & ACC & MCC & ACC & MCC & ACC & MCC & ACC & MCC & ACC & MCC & ACC & MCC\\
\midrule
LLaVA 1.5-7B 1-shot&35.60 & 17.73 & 20.20 & 0.94 & 41.80 & 26.70 & 55.36 & 41.83 & 54.60 & 39.33 & 58.79 & 44.27 & 44.39 & 28.46\\
LLaVA 1.5-7B 2-shot&10.60 & 5.68 & 8.59 & 5.72 & 14.20 & 11.59 & 18.20 & 16.18 & 46.20 & 34.57 & 49.61 & 35.99 & 24.57 & 18.29\\
InternVL 3-8B 1-shot&38.00 & 17.64 & 29.04 & 5.55 & 45.40 & 27.36 & 55.36 & 40.53 & 38.20 & 17.96 & 53.65 & 37.40 &  43.28 & 24.41\\
InternVL 3-8B 2-shot&42.60 & 23.74 & 30.30 & 6.82 & 43.40 & 24.72 & 56.86 & 42.46 & 42.60 & 23.92 & 53.50 & 36.94 & 44.88 & 26.43\\
lingshu-7B 1-shot& 38.80 & 19.01 & 34.85 & 14.62 & 49.80 & 33.25 & 69.08 & 58.83 & 49.20 & 33.08 & 65.01 & 51.82  & 51.12 & 35.10\\
lingshu-7B 2-shot& 45.60 & 27.99 & 38.13 & 17.71 & 48.40 & 31.26 & 70.32 & 60.57 & 51.00 & 34.97 & 68.58 & 56.05 & 53.67 & 38.09\\
Hulu-Med-7B 1-shot& 43.40 & 24.78 & 30.81 & 7.86 & 57.60 & 43.66 & 65.84 & 54.53 & 49.20 & 32.28 & 60.19 & 47.34 & 51.17 & 35.07\\
Hulu-Med-7B 2-shot& 49.60 & 33.02 & 34.09 & 12.33 & 59.60 & 46.31 & 68.58 & 58.59 & 49.40 & 32.60 & 63.61 & 50.78 & 54.15 & 38.94\\
\rowcolor{blue!10} \Method-Qwen3-8B &  \textbf{76.60} & \textbf{68.77} & \textbf{92.42} & \textbf{89.93} & \textbf{82.80} & \textbf{77.05} & \textbf{93.52} & \textbf{91.39} & \textbf{75.20} & \textbf{66.89} & \textbf{72.32} & \textbf{61.28} & \textbf{82.14} & \textbf{75.88}\\
\bottomrule
\end{tabular}
}
\end{table}

\begin{table}[ht]
\centering
\tiny
\setlength{\tabcolsep}{4.2pt}
\renewcommand{\arraystretch}{0.9}
\caption{Comparison of \Method with few-shot baselines under \textit{Open-QA} and \textit{report generation} settings.}
\vspace{2mm}
\label{tab:few-shot-open}
\resizebox{1\columnwidth}{!}{
\begin{tabular}{lcccccccccc cccccc cccc}
\toprule
\multirow{3}{*}{Model} & \multicolumn{10}{c}{In-distribution Task} & \multicolumn{6}{c}{Out-of-distribution Task} & \multicolumn{4}{c}{\multirow{1}{*}{Avg.}} \\
\cmidrule(lr){2-11}\cmidrule(lr){12-17}\cmidrule(lr){18-21}
&\multicolumn{2}{c}{Admissions} &
\multicolumn{2}{c}{IE} &
\multicolumn{2}{c}{POE} &
\multicolumn{2}{c}{Services} &
\multicolumn{2}{c}{MIMIC-CXR} &
\multicolumn{2}{c}{Transfers} &
\multicolumn{2}{c}{Obesity} &
\multicolumn{2}{c}{PadChest} &
\multicolumn{2}{c}{\multirow{1}{*}{Open-QA}} &
\multicolumn{2}{c}{\multirow{1}{*}{Report}}
\\
\cmidrule(lr){2-3}\cmidrule(lr){4-5}\cmidrule(lr){6-7}\cmidrule(lr){8-9}\cmidrule(lr){10-11}\cmidrule(lr){12-13}\cmidrule(lr){14-15}\cmidrule(lr){16-17}\cmidrule(lr){18-19}\cmidrule(lr){20-21}
& F1-Bio & BLEU-1 & F1-Bio & BLEU-1 & F1-Bio & BLEU-1 & F1-Bio & BLEU-1 & F1-Rad & F1-Bio & F1-Bio & BLEU-1 & F1-Bio & BLEU-1& F1-Rad & F1-Bio& F1-Bio & BLEU-1 &F1-Rad & F1-Bio\\
\midrule
LLaVA 1.5-7B 1-shot&72.42 & 23.91 & 68.45 & 17.87 & 71.84 & 22.52 & 72.93 & 22.99 & 4.17 & 68.19 & 76.07 & 28.96 & 73.11 & 11.48 & 9.25 & 67.83 & 72.47 & 21.29 & 6.71 & 68.01\\
LLaVA 1.5-7B 2-shot&51.98 & 6.14 & 50.39 & 3.98 & 51.28 & 5.24 & 52.83 & 6.56 & 2.44 & 54.40 & 68.85 & 23.58 & 67.65 & 9.44 & 4.69 & 45.58  & 57.16 & 9.16 & 3.57 & 49.99\\
InternVL 3-8B 1-shot&74.40 & 27.94 & 75.70 & 28.50 & 74.89 & 28.06 & 74.92 & 28.32 & 9.18 & 67.77 & 75.44 & 30.32 & 69.46 & 7.04 & 8.63 & 63.21 & 74.13 & 25.03 & 8.91 & 65.49\\
InternVL 3-8B 2-shot&75.48 & 30.00 & 76.24 & 30.14 & 75.69 & 30.34 & 75.61 & 30.27 & 9.32 & 68.25 & 76.11 & 31.96 & 69.45 & 6.98 & 8.31 & 63.76 & 74.76 & 26.62 & 8.82 & 66.01\\
lingshu-7B 1-shot& 77.09 & 38.61 & 75.65 & 30.27 & 76.54 & 33.60 & 76.76 & 34.11 & 11.98 & 75.19 & 76.97 & 32.17 & 73.04 & 10.67 & 7.85 & 67.36 &  76.01 & 29.91 & 9.92 & 71.27\\
lingshu-7B 2-shot& 77.55 & 39.27 & 76.18 & 30.57 & 76.74 & 34.79 & 77.21 & 34.89 & 11.53 & 74.36 & 77.25 & 33.63 & 72.88 & 10.54 & 7.67 & 66.75 & 76.30 & 30.61 & 9.60 & 70.55\\
Hulu-Med-7B 1-shot& 76.93 & 31.16 & 75.42 & 24.08 & 76.10 & 29.40 & 76.56 & 29.89 & 13.86 & 75.77 & 76.91 & 26.78 & 72.87 & 10.58 & 8.86 & 64.90 & 75.80 & 25.32 & 11.36 & 70.34\\
Hulu-Med-7B 2-shot& 77.04 & 32.19 & 75.83 & 23.07 & 76.25 & 30.06 & 76.76 & 29.04 & 14.50 & 76.32 & 77.01 & 29.95 & 72.85 & 10.36 & 8.29 & 65.52 & 75.96 & 25.78 & 11.40 & 70.92\\
\rowcolor{blue!10} \Method-Qwen3-8B & \textbf{80.82} & \textbf{52.45} & \textbf{82.21} & \textbf{51.69} & \textbf{81.02} & \textbf{51.97} & \textbf{80.21} & \textbf{49.19} & \textbf{17.49} & \textbf{79.03} & \textbf{79.57} & \textbf{46.42} & \textbf{75.37} & \textbf{15.65} & \textbf{25.52} & \textbf{73.93}  & \textbf{79.87} & \textbf{44.56} & \textbf{21.51} & \textbf{76.48}\\
\bottomrule
\end{tabular}
}
\end{table}

\subsection{Hyperparameter Study}
\label{sec:appendix-hyperparameter}
\begin{figure}[h]
\centering

\begin{subfigure}{0.32\textwidth}
    \centering
    \includegraphics[width=\linewidth]{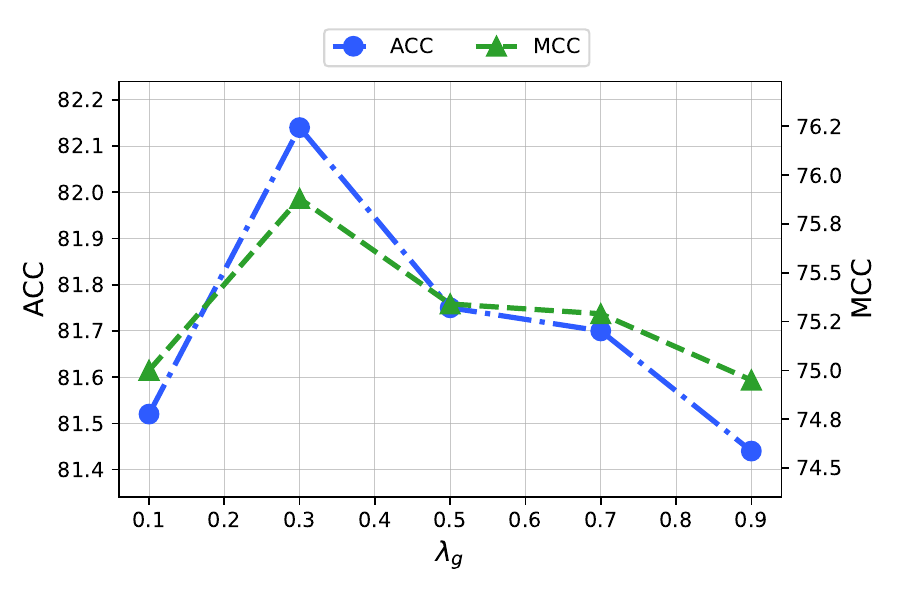}
    \caption{Hyperparameter study of $\lambda_g$ under \textit{Closed-QA} setting.}
\end{subfigure}
\hfill
\begin{subfigure}{0.32\textwidth}
    \centering
    \includegraphics[width=\linewidth]{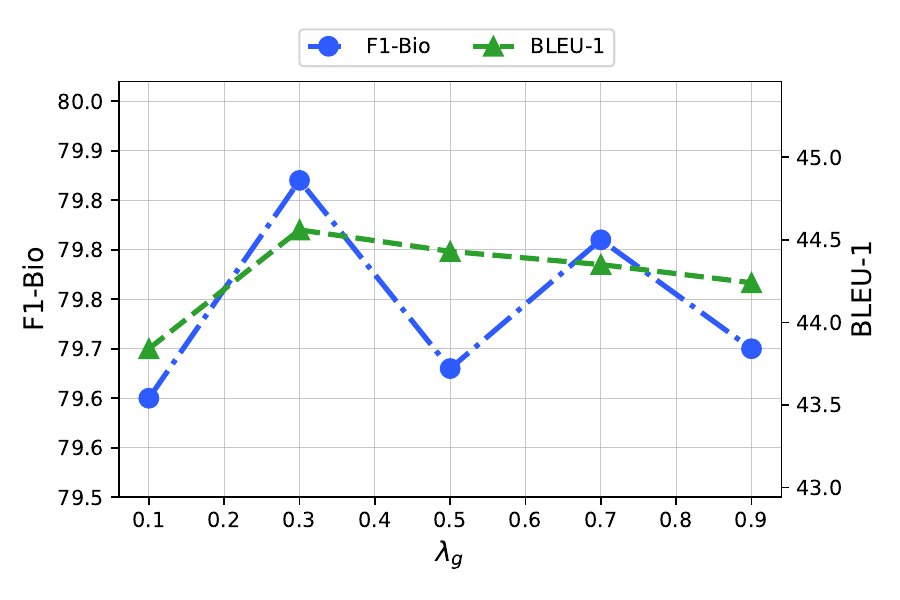}
    \caption{Hyperparameter study of $\lambda_g$ under \textit{Open-QA} setting.}
\end{subfigure}
\hfill
\begin{subfigure}{0.32\textwidth}
    \centering
    \includegraphics[width=\linewidth]{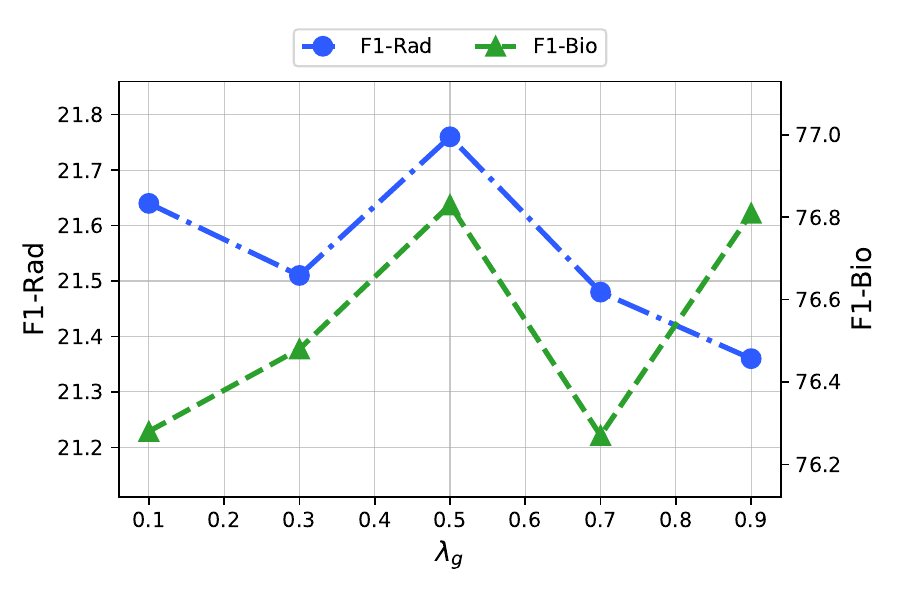}
    \caption{Hyperparameter study of $\lambda_g$ under \textit{Report Generation} setting.}
\end{subfigure}

\vspace{2mm}

\begin{subfigure}{0.32\textwidth}
    \centering
    \includegraphics[width=\linewidth]{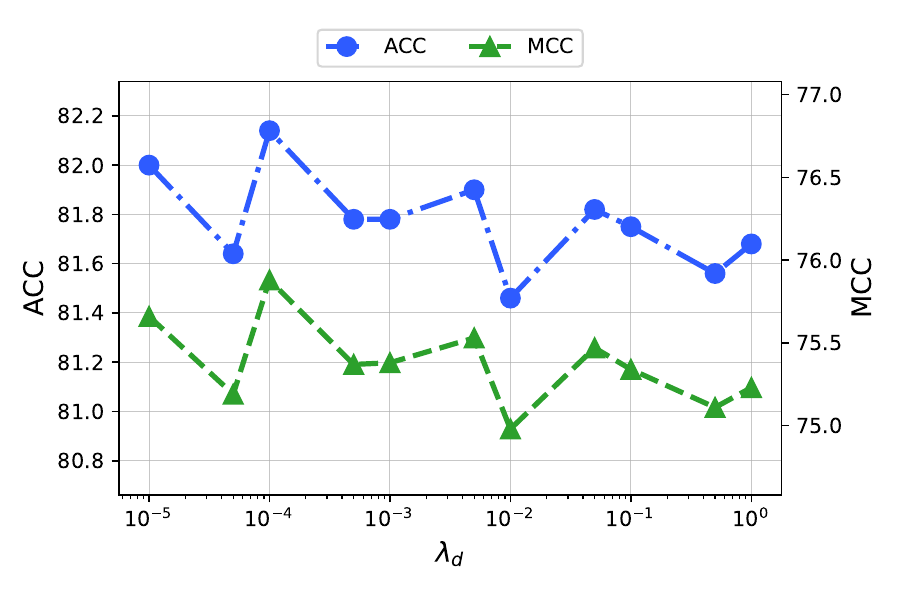}
    \caption{Hyperparameter study of $\lambda_d$ under \textit{Closed-QA} setting.}
\end{subfigure}
\hfill
\begin{subfigure}{0.32\textwidth}
    \centering
    \includegraphics[width=\linewidth]{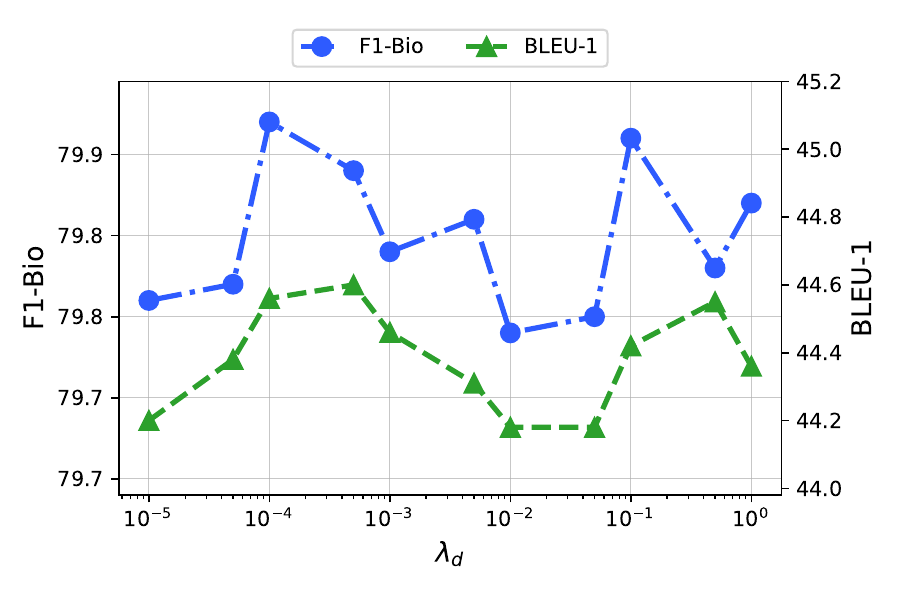}
    \caption{Hyperparameter study of $\lambda_d$ under \textit{Open-QA} setting.}
\end{subfigure}
\hfill
\begin{subfigure}{0.32\textwidth}
    \centering
    \includegraphics[width=\linewidth]{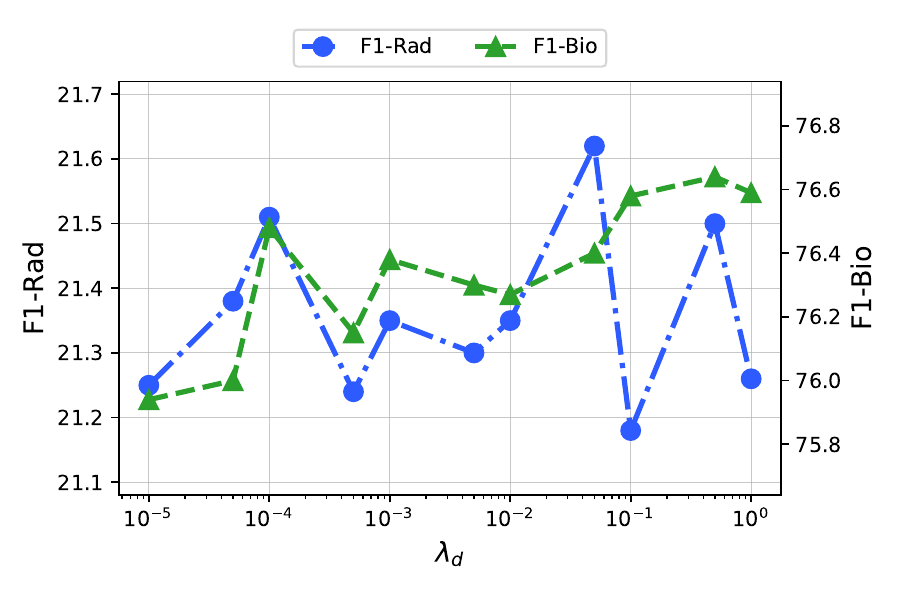}
    \caption{Hyperparameter study of $\lambda_d$ under \textit{Report Generation} setting.}
\end{subfigure}

\caption{Hyperparameter study on $\lambda_g$ and $\lambda_d$.}
\label{fig:hyper}
\end{figure}

We conduct a sensitivity study on two key hyperparameters, $\lambda_g$ and $\lambda_d$, with results shown in Figure~\ref{fig:hyper}. 
As observed in Figure~\ref{fig:hyper}(a–c), performance consistently peaks at moderate values of $\lambda_g$ (around 0.3–0.5), 
highlighting the importance of balancing the generation objective and cohort-aware signals. 
Both smaller and larger values lead to performance degradation, corresponding to the under- and over-dominance of the generation objective, respectively.

In contrast, Figure~\ref{fig:hyper}(d–f) shows that the model is relatively insensitive to $\lambda_d$, with stable performance across several orders of magnitude. 
The best results are typically achieved around $10^{-4}$ or $10^{-1}$, suggesting that the drift loss primarily acts as a regularizer.

Overall, these results demonstrate that our method achieves robust performance without requiring precise hyperparameter tuning.

\newpage
\section{Case Study}
\label{sec:appendix-case}


\begin{figure}[H]
    \centering
    \includegraphics[width=1\linewidth]{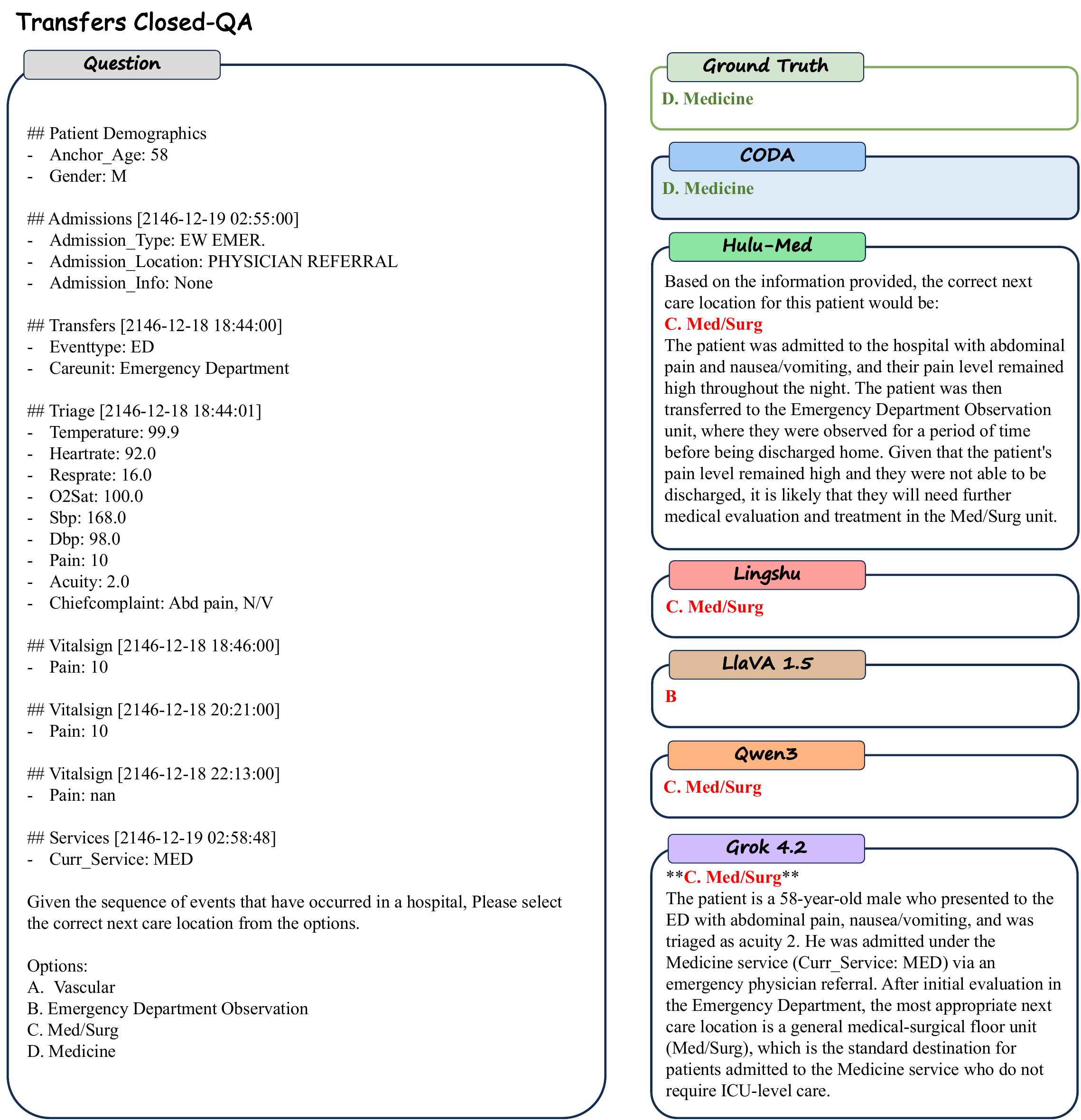}
    \caption{A case study of Transfers task under \textit{Closed-QA} setting.}
    \label{fig:placeholder}
\end{figure}

\newpage
\begin{figure}[H]
    \centering
    \includegraphics[width=1\linewidth]{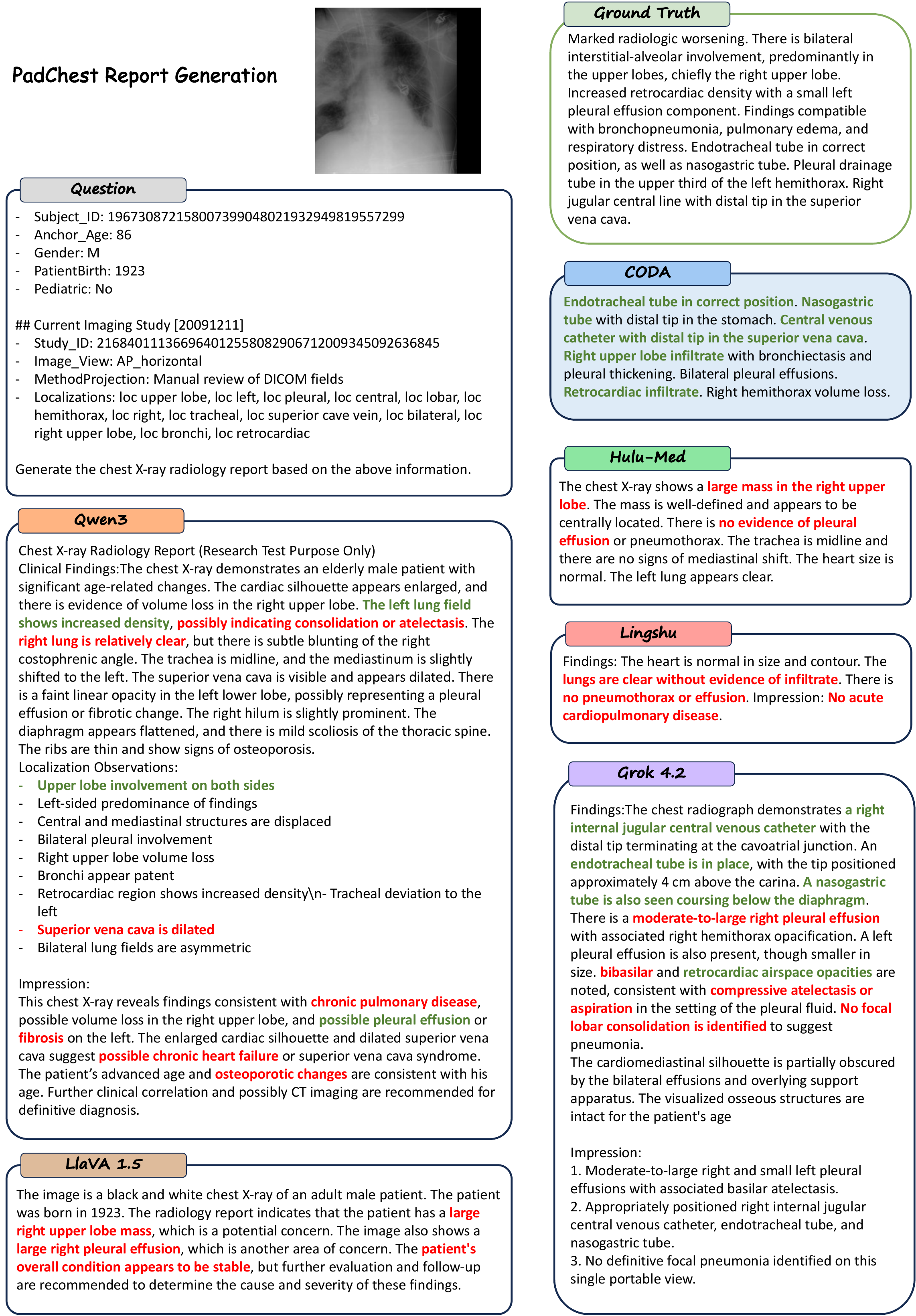}
    \caption{A case study of PadChest dataset under \textit{Report Generation} setting.}
    \label{fig:placeholder}
\end{figure}
\newpage

\section{Broader Impacts}
\label{sec:appendix-impact}
Current medical foundation models overlook patient-specific characteristics, which are critical for accurate clinical decision-making. By explicitly modeling cohort structure and individual-level deviations, \Method has the potential to support clinicians in delivering more personalized care, particularly in settings with heterogeneous patient populations, and to enhance transparency by exposing cohort-level patterns and patient-specific factors that influence model predictions.

At the same time, several risks should be considered. First, cohort construction may inadvertently encode biases present in the underlying data, potentially leading to disparate performance across demographic groups. Second, reliance on structured features and summary statistics may limit applicability in scenarios where such information is incomplete or noisy. Third, as with other generative models, there remains a risk of producing plausible but incorrect outputs, which could be misinterpreted in high-stakes clinical contexts.

To mitigate these risks, future work should incorporate fairness-aware cohort construction, uncertainty estimation, and human-in-the-loop validation to ensure safe deployment. In addition, rigorous evaluation across diverse populations and clinical settings is necessary to assess generalizability and prevent unintended harms. Overall, while \Method offers a step toward more interpretable and patient-aware medical AI, careful consideration of these societal implications is essential for responsible use.



\end{document}